\documentclass[journal]{IEEEtran}

\usepackage{amsmath}
\usepackage{graphicx}
\usepackage{algorithm,float} 
\usepackage{algpseudocode}
\usepackage{caption,subcaption}
\usepackage{amsmath,amsfonts,amsthm,mdwmath} 
\usepackage{graphics,graphicx}	
\DeclareGraphicsExtensions{.eps,.pdf,.gif,.png,.jpg}
\usepackage{url}
\usepackage{algorithm}
\usepackage{multicol}
\usepackage{color}
\usepackage{booktabs}
\usepackage[draft=true]{hyperref}
\newcommand\norm[1]{\left\lVert#1\right\rVert}

\begin{document}
	\title{A Crowdsourcing Framework for On-Device Federated Learning }
	\author{Shashi~Raj~Pandey,~\IEEEmembership{Student~Member,~IEEE,~}
		Nguyen~H.~Tran,~\IEEEmembership{Senior~Member,~IEEE,~}
		Mehdi~Bennis,~\IEEEmembership{Senior~Member,~IEEE,~}
		Yan~Kyaw~Tun, Aunas~Manzoor,~and~Choong~Seon~Hong,~\IEEEmembership{Senior~Member,~IEEE}
		\thanks{Manuscript received May 19, 2019; revised September 7, 2019, December 17, 2019 and January 13, 2020; accepted January 28, 2020. Date of publication......; date of current version ....This work was supported by Institute of Information \& communications Technology Planning \& Evaluation (IITP) grant funded by the Korea government (MSIT) (No.2019-0-01287, Evolvable Deep Learning Model Generation Platform for Edge Computing) and the National Research Foundation of Korea(NRF) grant funded by the Korea government (MSIT) (NRF-2017R1A2A2A05000995). A preliminary version of this work has been presented at IEEE GLOBECOM 2019 \cite{shashi}. \textit{(Corresponding author:
			Choong Seon Hong.)}}
		\thanks{Shashi Raj Pandey, Yan Kyaw Tun, Aunas Manzoor, and Choong Seon Hong  are with the Department of Computer Science and Engineering, Kyung Hee University,  Yongin-si, Gyeonggi-do 17104, Rep. of Korea, e-mail: {\{shashiraj, ykyawtun7, aunasmanzoor, cshong\}@khu.ac.kr}.}
		\thanks{Nguyen H. Tran is with the School of Computer Science, The University of Sydney, NSW 2006, Australia, and also with the Department of
			Computer Science and Engineering, Kyung Hee University, Seoul 17104,
			South Korea (email: {nguyen.tran@sydney.edu.au}).}
		\thanks{Mehdi Bennis is with the Center for Wireless Communications, University of Oulu, 90014 Oulu, Finland, and also with the Department of
			Computer Science and Engineering, Kyung Hee University, Seoul 17104,
			South Korea (email: {mehdi.bennis@oulu.fi}).}
		\thanks{}}    	
	\maketitle
	\begin{abstract}
		Federated learning (FL) rests on the notion of training a global model in a decentralized manner. Under this setting, mobile devices perform computations on their local data before uploading the required updates to improve the global model. However, when the participating clients implement an uncoordinated computation strategy, the difficulty is to handle the communication efficiency (i.e., the number of communications per iteration) while exchanging the model parameters during aggregation. Therefore, a key challenge in FL is how users participate to build a high-quality global model with communication efficiency. We tackle this issue by formulating a utility maximization problem, and propose a novel crowdsourcing framework to leverage FL that considers the communication efficiency during parameters exchange. First, we show an incentive-based interaction between the crowdsourcing platform and the participating client's independent strategies for training a global learning model, where each side maximizes its own benefit. We formulate a two-stage Stackelberg game to analyze such scenario and find the game's equilibria. Second, we formalize an admission control scheme for participating clients to ensure a level of local accuracy. Simulated results demonstrate the efficacy of our proposed solution with up to 22$\%$ gain in the offered reward.
	\end{abstract}

	\begin{IEEEkeywords}
		Decentralized machine learning, federated learning (FL), mobile crowdsourcing, incentive mechanism, Stackelberg game. 
	\end{IEEEkeywords} 
	\IEEEpeerreviewmaketitle
	
	\section{Introduction}

	\subsection{Background and motivation}

	Recent years have admittedly witnessed a tremendous growth in the use of Machine Learning (ML) techniques and its applications in  mobile devices. On one hand, according to International Data Corporation, the shipments of smartphones reached 3 billions in 2018 \cite{idc}, which implies  a large crowd of mobile users generating personalized data via interaction with mobile applications, or with the use of in-built sensors (e.g., cameras, microphones and GPS) exploited efficiently by mobile crowdsensing paradigm (e.g., for indoor localization, traffic monitoring, navigation \cite{ganti2011mobile}, \cite{zhang2013robust}, \cite{wu2015smartphones}, \cite{koukoumidis2011signalguru}). On the other hand, mobile devices are getting empowered extensively with specialized hardware architectures and computing engines such as the CPU, GPU and DSP (e.g., energy efficient Qualcomm Hexagon Vector eXtensions on Snapdragon 835 \cite{a}) for solving diverse machine learning problems. Gartner predicts that 80 percent of smartphones will have on-device AI capabilities by 2022. With dedicated chipsets, it will empower smartphone makers to achieve market gain by offering more secured facial recognition system, the ability to understand user behaviors and offer predictive future \cite{b12}. This means \textit{on-device intelligence} will be ubiquitous!
	
	In the backdrop to these exciting possibilities with on-device intelligence, a White House report on principle of data minimization had been published in 2012 to advocate the privacy of consumer data \cite{house2012consumer}. The direct application of this is the ML technique that leaves the training data distributed on the mobile devices, called \textit{Federated Learning} \cite{a}, \cite{a3}, \cite{konevcny2016federated}, \cite{konecny2016federated}, \cite{mcmahan2017federated}. This technique unleashes a new collaborative ecosystem in ML to build a shared learning model while keeping the training data locally on user devices, which complies with the data minimization principle and protects user data privacy.  Unlike the conventional approaches of collecting all the training data in one place to train a learning model, the mobile users (participating clients) perform computation for the updates on their local training data with the current global model parameters, which are then aggregated and broadcasted back by the centralized coordinating server. This is an iterative process that undergoes until an accuracy level of the learning model is reached. By this way, FL decouples the training process to learn a global model by eliminating the mobility of local training data. 
	
	In another report, research organizations estimate that over 90\% of the data will be stored and processed locally \cite{kelly2016internet} (e.g., at the network edge), which provides an immense exposure to extract the benefits of FL. Also, because of the huge market potential of the untapped private data, FL is a promising tool to exploit more personalized service oriented applications. 
	
	Local computations at the devices and their communication with the centralized coordinating server are interleaved in a complex manner to build a global learning model. Therefore, a communication-efficient FL framework \cite{konecny2016federated}, \cite{mcmahan2016communication} requires solving several challenges. Furthermore, because of limited data per device to train a high-quality learning model, the difficulty is to incentivize a large number of mobile users to ensure cooperation. This important aspect in FL has been overlooked so far, where the question is \textit{how can we motivate a number of participating clients, collectively providing a large number of data samples to enable FL without sharing their private data?} Note that, both participating clients and the server can benefit from training a global model. However, to fully reap the benefits of high-quality updates, the multi-access edge computing (MEC) server has to incentivize clients for participation. In particular, under heterogeneous scenarios, such as an adaptive and cognitive-communication network, client's participation in FL can spur collaboration and provide benefits for operators to accelerate and deliver network-wide services \cite{wang2018edge}. Similarly, clients in general are not concerned with the reliability and scalability issues of FL \cite{kim2018device}. Therefore, to incentivize users to participate in the collaborative training, we require a market place. For this purpose, we present a value-based compensation mechanism to the participating clients, such as a bounty (e.g., data discount package), as per their level of participation  in the crowdsourcing framework. This is reflected in terms of local accuracy  level, i.e., \textit{quality of solution to the local subproblem}, in which the framework will protect the model from imperfect updates by restricting the clients trying to compromise the model (for instance, with skewed data because of its i.i.d nature or data poisoning) \cite{ganti2011mobile}. Moreover, we cast the global loss minimization problem as a primal-dual optimization problem, instead of adopting traditional gradient descent learning algorithm in the federated learning setting (e.g., FedAvg \cite{mcmahan2016communication}). This enables in (a) proper assessment of the quality of the local solution to improve personalization and fairness amongst the participating clients while training a global model, (b) effective decoupling of the local solvers, thereby balancing communication and computation in the distributed setting.

	The goal of this paper is two-fold: First, we formalize an incentive mechanism to develop a participatory framework for mobile clients to perform FL for improving the global model. Second, we address the challenge of maintaining communication efficiency while exchanging the model parameters with a number of participating clients during aggregation. Specifically, communication efficiency in this scenario accounts for communications per iteration with an arbitrary algorithm to maintain an acceptable accuracy level for the global model.
	\subsection{Contributions}
	In this work, we design and analyze a novel crowdsourcing framework to realize the FL vision. Specifically, our contributions are summarized as follows: 
	\begin{itemize}
		\item\textbf{A crowdsourcing framework to enable communication -efficient FL}. We design a crowdsourcing framework, in which FL participating clients iteratively solve the local learning subproblems for an accuracy level subject to an offered incentive. We then establish a communication-efficient cost model for the participating clients. We then formulate an incentive mechanism to induce the necessary interaction between the MEC server and the participating clients for the FL in Section \ref{sec:Incentive_Mechanism}.
		\item \textbf{ Solution approach using Stackelberg game}.
		With the offered incentive, the participating clients independently choose their strategies to solve the local subproblem for a certain accuracy level in order to minimize their participation costs. Correspondingly, the MEC server builds a high quality centralized model characterized by its utility function, with the data distributed over the participating clients by offering the reward. 
		We exploit this tightly coupled motives of the participating clients and the MEC server as a two-stage Stackelberg game. The equivalent optimization problem is characterized as a mixed-boolean programming which requires an exponential complexity effort for finding the solution. We analyze the game's equilibria and propose a linear complexity algorithm to obtain the optimal solution.
		\item \textbf{Participant's response analysis and case study}. We next analyze the response behavior of the participating clients via the solutions of the Stackelberg game, and establish the efficacy of our proposed framework via case studies. We show that the linear-complexity solution approach attains the same performance as the mixed-boolean programming problem. Furthermore, we show that our mechanism design can achieve the optimal solution while outperforming a heuristic approach for attaining the maximal utility with up to $22\%$ of gain in the offered reward.
		\item \textbf{Admission control strategy}. Finally, we show that it is significant to have certain participating clients to guarantee the communication efficiency for an accuracy level in FL. We formulate a probabilistic model for threshold accuracy estimation and find the corresponding number of participation required to build a high-quality learning model. We analyze the impact of the number of participation in FL while determining the threshold accuracy level with closed-form solutions. Finally, with numerical results we demonstrate the structure of admission control model for different configurations. 
	\end{itemize}
	The remainder of this paper is organized as follows. We  review related work in Section \ref{sec:Related_works}, and present the system model in Section \ref{sec:SystemModel}. In Section \ref{sec:Incentive_Mechanism}, we formulate an incentive mechanism with a two-stage Stackelberg game, and investigate the Nash equilibrium of the game with simulation results in Section \ref{sim:Simulation_a}. An admission control strategy is formulated to define a minimum local accuracy level, and numerical analysis is presented in Section \ref{sec:Admission_Control}. Finally, conclusions are drawn in Section \ref{sec:Conclusions}.

	\section{Related Work} \label{sec:Related_works}
	
	The  unprecedented amount of data necessitates the use of distributed computational framework to provide solutions for various machine learning applications \cite{konevcny2016federated}--\cite{mcmahan2016communication}. Using distributed optimization techniques, researches on decentralized machine learning largely focused on competitive algorithms to train learning models the number of cluster nodes \cite{ma2017distributed}, \cite{shamir2014distributed}, \cite{bao2018online}, \cite{iandola2016firecaffe}, with balanced and i.i.d data.  
	
	Setting a different motivation, FL recently has attracted an increasing interest \cite{a}, \cite{konevcny2016federated}, \cite{konecny2016federated}, \cite{mcmahan2017federated}, \cite{mcmahan2016communication}, \cite{nyg} in which collaboration of the number of devices with non-i.i.d and unbalanced data is adapted  to train a learning model. In the pioneering works \cite{konevcny2016federated}, \cite{konecny2016federated}, the authors presented the setting for federated optimization, and related technical challenges to understand the convergence properties in FL. Existing work studied these issues. For example, Wang, Shiqiang, et al. \cite{wang2018edge} theoretically analyzed the convergence rate of the distributed gradient descent. In this detailed work, the authors focus on deducing the optimal global aggregation frequency in a distributed learning setting to minimize the loss function of the global problem. Their problem considers resource constrained edge computing system. However, the setting differs with our proposed model where we have introduced the notion of participation, and proposed a game theoretic interaction between the workers (participating clients) and the master (MEC server) to attain a cost effective FL framework. Earlier to this work,  McMahan, H. Brendan, et al. in \cite{mcmahan2016communication} proposed a practical variant of FL where the global aggregation was synchronous with a fixed frequency. The authors confirmed the effectiveness of this approach using various datasets. Furthermore, authors in \cite{ma2017distributed} extended the theoretical training convergence analysis results of \cite{mcmahan2016communication} to general classes of distributed learning approaches with communication and computation cost. For the deep learning architecture where the objectives are non-convex, authors in \cite{fedopthet} proposed an algorithm namely FedProx, a special case of FedAvg where a surrogate of the global objective function was used to efficiently ensure the empirical performance bound in FL setting. In this work, the authors demonstrated the improvement in performance as in their theoretical assumptions, both  in terms of robustness and convergence through a set of experiments.
	
	Recent works adapt and extend the core concepts in \cite{konevcny2016federated}, \cite{konecny2016federated}, \cite{mcmahan2016communication}  to develop a communication-efficient FL algorithm, where each participating clients in the federated learning setting independently computes their
		local updates on the current model and communicates with a central server to aggregate the parameters for the computation
		of a global model. The framework uses Federated Averaging (FedAvg) algorithm to reduce communication costs. In
		these regard, to characterize the communication and computation trade-off during model updates, distributed machine
		learning based on gradient descent is widely used. In the mentioned work \cite{konevcny2016federated}, a variant of distributed stochastic gradient
		descent (SGD) was used to attain parallelism and improved computation. Similarly, in \cite{konecny2016federated}, the authors discussed
		about a family of new randomized methods combining SGD, with primal and dual variants such as Stochastic Variance
		Reduced Gradient (SVRG), Federated Stochastic Variance Reduced Gradient (FSVRG) and Stochastic Dual Coordinate
		Ascent (SDCA). Further, in \cite{lin2017deep} the authors explained about the redundancy in gradient exchanges in distributed SGD,
		and proposed a Deep Gradient Compression (DGC) algorithm to enhance communication efficiency in FL setting. The
		performance of parallel SGD and mini-batch parallel SGD had been discussed in \cite{stich2018local}, \cite{fedopthet} for fast convergence and
		effective communication rounds. However, authors in their recent work \cite{stich2018local} argue for the sufficient improvement in
		generalization performance with the variant of local SGD rather than the large mini-batch sizes, even in a non-convex setting.
		In \cite{shamir2014communication}, the authors proposed the Distributed Approximate Newton (DANE) algorithm for precisely solving a general
		subproblem available locally before averaging their solutions. In the recent work \cite{wang2018adaptive}, the authors designed a
		robust method which applies the proposed periodic-averaging SGD (PASGD) technique to prevent communication delay
		in the distributed SGD setting. The idea in this work was to adapt the communication period such that it minimizes the
		optimization error at each wall-clock time. To this end, interestingly, in some of the latest works such as \cite{wang2019beyond}, the authors have well-studied and demonstrated the privacy risk scenario under collaborated learning mechanism such as FL.
	
	In contrast to the above research that has overlooked the participatory method to build a high-quality central ML model and its criticality, and primarily focused on the convergence of learning time with variants of learning algorithms, our work addresses the challenge in designing a communication and computational cost effective FL framework by exploring a crowdsourcing structure. In this regard, few recent studies have discussed about the participation to build a global ML model with FL as in  \cite{samarakoon2018distributed}, \cite{zhu2018towards}. Basically, in \cite{samarakoon2018distributed} the authors proposed a novel distributed approach based on FL to learn the network-wide queue dynamics in vehicular networks for achieving ultra-reliable low-latency communication (URLLC) via a joint power and resource allocation problem. The vehicles participate in FL to provide information related to sample events (i.e., queue lengths) to parameterize the distribution of extremes. In \cite{zhu2018towards}, the authors provided new design principles to characterize edge-learning and highlighted important research opportunities and applications with the new philosophy for wireless communication  called \textit{learning-driven communication}. The authors also presented some of the significant case studies and demonstrated the effectiveness of design principles in this regards. Further, recent work \cite{kim2018device} studied the block-chained FL architecture proposing the data reward and mining reward mechanism  for FL. However, these works largely provide a latency analysis for the related applications. Our paper focuses on the Stackelberg game-based incentive mechanism design to \textit{reveal the iteration strategy} of the participating clients by solving the local subproblems for building a high-quality centralized learning model. Interestingly, incentive mechanism has been studied for years in mobile crowdsourcing/crowdsensing systems, especially with auction mechanisms (e.g., \cite{zhang2016toward}, \cite{wei2015truthful}, \cite{jin2018incentive}), contract and tournament models (e.g, \cite{zhang2017incentive}, \cite{wen2015quality}) and Stackelberg game-based incentive mechanisms such as in \cite{duan2014motivating} and \cite{yang2012crowdsourcing}. However, the design goals were specific towards fair and truthful data trading of distributed sensing tasks. In this regard, the novelty of our model is that we untangle and analyze the complex interaction scenario between the participating clients and the aggregating edge server in the crowdsourcing framework to obtain a cost-effective global learning model without sharing local datasets. Moreover, the proposed incentive mechanism models such \textit{interactions} to enable \textit{communication-efficient} FL, which is able to achieve a target accuracy, in consideration with the performance metrics. Further, we adopt the dual formulation of the learning problem to better decompose the global problem into distributed subproblems for federated computation across the participating clients.

	\section{System Model}\label{sec:SystemModel}
	
\begin{figure}[t!]
		\centering
		\includegraphics[width=3in]{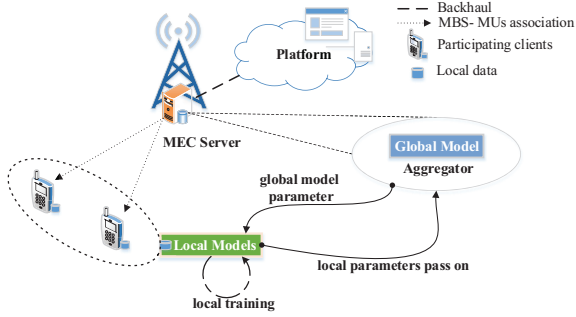}
		\caption{Crowdsourcing framework for decentralized machine learning.}
		\label{sysmodel}
	\end{figure}

	Fig. \ref{sysmodel} illustrates our proposed system model for the crowdsourcing framework to enable FL. The model consists of a number of mobile clients associated with a base station having a central coordinating server (MEC server), acting as a central entity. The server facilitates the computation of the parameters aggregation, and feedback the global model updates in each global iteration. We consider a set of participating clients $\mathcal{K} = \{1,2,\ldots,K \}$ in the crowdsourcing framework. The crowdsourcer (platform) can interact with mobile clients via an application interface, and aims at leveraging FL to build a global ML model. As an example, consider a case where the crowdsourcer (referred to as MEC server hereafter, to avoid any confusion) wants to build a ML model. Instead of just relying on available local data to train the global model at the MEC server, the global model is constructed utilizing the local training data available across several distributed mobile clients. Here, the global model parameter is first shared by the MEC server to train the local models in each participating client. The local model's parameters minimizing local loss functions are then sent back as feedback, and are aggregated to update the global model parameter. The process continues iteratively, until convergence.

	\subsection{Federated Learning Background} 

	For FL, we consider unevenly partitioned training data over a large number of participating clients to train the local models under any arbitrary learning algorithm. Each client $k$  stores its local dataset $\mathcal{D}_k $ of size $D_k$  respectively. Then, we define the training data size $D = \sum_{k=1}^{K}D_{k}$. 
	In a typical supervised learning setting, $\mathcal{D}_k$ defines the collection of data samples given as a set of  input-output pairs $\{x_i, y_i\}_{i=1}^{D_k}$, where $x_i \in \mathbb{R}^d$ is an input sample vector with $d$ features, and $y_i \in \mathbb{R}$ is the labeled output value for the sample $x_i$. The learning problem, for an input sample vector $x_i$ (e.g., the pixels of an image) is to find the  \textit{model parameter vector} $w\in \mathbb{R}^d$ that characterizes the output $y_i$ (e.g., the labeled output of the image, such as the corresponding product names in a store) with the loss function $f_i(w)$. Some examples of loss functions include $f_i(w) = \frac{1}{2}(x_i^Tw - y_i)^2, y_i \in \mathbb{R}$ for a linear regression problem and $f_i(w) = \max\{0, 1 -y_ix_i^Tw\},y_i \in \{-1,1\}$ for support vector machines. The term $x_i^Tw$ is often called a \textit{linear mapping function}.	
	 Therefore, the loss function based on the local data of client $k$, termed local subproblem is formulated as 
	\begin{equation}
	J_{k}(w) = \frac{1}{D_k} \sum \nolimits_{i = 1}^{D_k}f_i(w) + \lambda g(w),
	\label{eq:localloss}
	\end{equation}	
	where $w\in \mathbb{R}^d$ is the local model parameter, and $g(\cdot)$ is a regularizer function, commonly expressed as $g(\cdot) = \frac{1}{2}\norm{\cdot}^2$; $\forall \lambda \in [0,1]$. This characterizes the \textit{local model} in the FL setting.
	\begin{algorithm}[t!]
		\caption{Federated Learning Framework }\label{euclid}
		\begin{algorithmic}[1]
			\State \textbf{Input:} Initialize dual variable  $\alpha^0 \in \mathbb{R}^D$, $D_k, \forall k \in K$.
			\For {\text{each aggregation round}}
			\For {k $\in \mathcal{K}$}
			\State Solve local subproblems \eqref{eq:learning_problem_local} in parallel.
			\State Update local variables as in \eqref{eq:local_variable}. 
			\EndFor 
			\State Aggregate to update global parameter as in \eqref{eq:global_variable}.
			\EndFor					
		\end{algorithmic}
		\label{Algorithm1}		
	\end{algorithm}
	
	\textbf{\textit{Global Problem}}: At the MEC server, the global problem can be represented as the finite-sum objective of the form 
	\begin{equation}
	\underset{w \in \mathbb{R}^d}{\text{min}}J(w) \ \  \text{where} \ \ J(w) \equiv \frac{\sum\nolimits_{k = 1}^K D_kJ_k (w)}{D}.
	\label{eq:learning_problem}
	\end{equation}
	Problems of such structure as in \eqref{eq:learning_problem} where we aim to minimize an average of $K$ local objectives are well-known as \textit{distributed consensus problems} \cite{boyd2011distributed}.
	
	\textbf{\textit{Solution Framework under Federated Learning}:} We recast the regularized global problem in \eqref{eq:learning_problem} as
	\begin{equation}
	\underset{w \in \mathbb{R}^d}{\text{min}}J(w) := \frac{1}{D} \sum \nolimits_{i = 1}^{D}f_i(w) + \lambda g(w),
	\label{eq:learning_problem_central}
	\end{equation}
	and decompose it as a dual optimization problem\footnote{The duality gap provides a certificate to the quality of local solutions and facilitates distributed training.} in a distributed scenario \cite{shalev2014accelerated} amongst $K$ participating  clients. For this, at first, we define $X \in \mathbb{R}^{d \times D_k}$ as a matrix with columns having data points for $i \in \mathcal{D}_k, \forall k$. Then, the corresponding dual optimization problem of \eqref{eq:learning_problem_central} for a convex loss function $f$ is 
	\begin{equation}
	\underset{\alpha \in \mathbb{R}^{D}}{\text{max}} \boldsymbol{\mathcal{G}}(\alpha) := \frac{1}{D} \sum \nolimits_{i = 1}^{D} - f_i^*(- \alpha_i) - \lambda g^*(\phi(\alpha)),
	\label{eq:learning_problem_dual}
	\end{equation}	
	where  $\alpha \in \mathbb{R}^{D}$ is the dual variable mapping to the primal candidate vector, $f_i^*$ and $g^*$ are the convex conjugates of $f_i$ and $g$ respectively \cite{boyd2004convex}; $\phi(\alpha) = \frac{1}{\lambda D} X \alpha$.  With the optimal value of dual variable $\alpha^*$ in \eqref{eq:learning_problem_dual}, we have $w(\alpha^*) = \nabla g^*(\phi(\alpha^*)) $ as the optimal solution of \eqref{eq:learning_problem_central} \cite{shalev2014accelerated}. For the ease of representation, we will use $\phi \in \mathbb{R}^d$ for $\phi{(\alpha)}$ hereafter. We consider that $g$ is a strongly convex function, i.e., $g^*(\cdot)$ is continuous differentiable. Then, the solution is obtained following an iterative approach to attain a global accuracy $0 \le \epsilon \le 1$ (i.e., $\mathbb{E}\left[ \boldsymbol{\mathcal{G}}(\alpha) - \boldsymbol{\mathcal{G}}(\alpha^*)\right] < \epsilon$). 

	Under the distributed setting, we further define data partitioning notations for clients $k \in \mathcal{K}$ to represent the working principle of the framework. Let us define  a weight vector $\varrho_{[k]} \in \mathbb{R}^D$ at the local subproblem $k$ with its elements zero for the unavailable data points.
	Following the assumption of having $f_i$ as $(1 / \gamma)$-smooth and 1-strongly convex of $g$ to ensure convergence, its consequences is the approximate solution to the local problem $k$ defined by the dual variables $\alpha_{[k]}$, $\varrho_{[k]}$, characterized as
	\begin{equation}
	\underset{\varrho_{[k]}\in \mathbb{R}^{D}}{\text{max}} \boldsymbol{\mathcal{G}}_k (\varrho_{[k]}; \phi, \alpha_{[k]}),
	\label{eq:learning_problem_local}
	\end{equation}
	where $\boldsymbol{\mathcal{G}}_k (\varrho_{[k]}; \phi, \alpha_{[k]}) = -\frac{1}{K} - \langle \nabla (\lambda g^*(\phi(\alpha))), \varrho_{[k]} \rangle  - \frac{\lambda}{2} \Vert\frac{1}{\lambda D}X_{[k]}\varrho_{[k]}  \Vert^2$ is defined with a matrix $X_{[k]}$ columns having data points for $i \in \mathcal{D}_k$, and zero padded otherwise.
	Each participating client $k \in \mathcal{K}$ iterates over its computational resources using any arbitrary solver to solve its local problem \eqref{eq:learning_problem_local} with a local relative $\theta_k$ accuracy that characterizes the quality of the local solution, and produces a random output $\varrho_{[k]}$ satisfying
	\begin{equation}
	\mathbb{E}\left[ \boldsymbol{\mathcal{G}}_k(\varrho^*_{[k]}) -  \boldsymbol{\mathcal{G}}_k(\varrho_{[k]})  \right] \le \theta_k \left[  \boldsymbol{\mathcal{G}}_k(\varrho^*_{[k]}) -  \boldsymbol{\mathcal{G}}_k(0) \right]. 
	\end{equation}
	Note that, with local (relative) accuracy $\theta_{k} \in [0,1]$, the value of $\theta_{k}=1$ suggests that no improvement was made by the local solvers during successive local iterations. Then, the local dual variable is updated as follows:
	\begin{equation}
	\alpha^{t+1}_{[k]} := \alpha^t_{[k]} + \varrho^t_{[k]},\forall k \in \mathcal{K}.
	\label{eq:local_variable} 
	\end{equation}
	Correspondingly, each participating client will broadcast the local parameter defined as $\Delta \phi^t_{[k]}:= \frac{1}{\lambda D} X_{[k]}\varrho^t_{[k]} $,  during each round of communication to the MEC server. The MEC server aggregates the local parameter (averaging) with the following rule:
	\begin{equation}
	\phi^{t+1} := \phi^t + \frac{1}{K}\sum \nolimits_{k = 1}^{K} \Delta \phi^t_{[k]},
	\label{eq:global_variable}
	\end{equation}
	and distributes the global change in $\phi$ to the participating clients, which is used to solve \eqref{eq:learning_problem_local} in the next round of local iterations. This way we observe the decoupling of global model parameter from the need of local clients' data\footnote{Note that we consider the availability of quality of data with each participating client for solving a corresponding local subproblem. Further related demonstration on dependency of the normalized data size and accuracy can be found in \cite{niyato2016market}.} for training a global model.

	Algorithm \ref{Algorithm1} briefly summarizes the FL framework as an iterative process to solve the global problem characterized in \eqref{eq:learning_problem_central} for a global accuracy level. The iterative process (S2)-(S8) of Algorithm \ref{Algorithm1} terminates when the global accuracy $\epsilon$ is reached. A participating client $k$ strategically\footnote{Fewer iterations might not be sufficient to have an optimal local solution \cite{wang2018edge}.} iterates over its local training data $\mathcal{D}_k$ to solve the local subproblem \eqref{eq:learning_problem_local} up to an accuracy $\theta_k$. In each communication round with the MEC server, the participating clients \textit{synchronously} pass on their parameters $\Delta\phi_{[k]}$ using a shared wireless channel. The MEC server then aggregates the local model parameters $\phi$ as in \eqref{eq:global_variable}, and broadcasts the global parameters required for the participating clients to solve their local subproblems for the next communication round. 
	\begin{figure}[t!]
		\centering
		\includegraphics[width=3in]{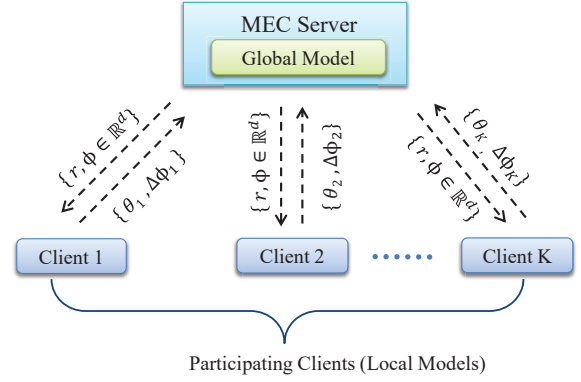}
		\caption{Interaction environment of federated learning setting under crowdsourcing framework.}
		\label{fig:message_pass}
	\end{figure}
	Within the framework, consider that each participating client uses any arbitrary optimization algorithm (such as \textit{Stochastic Gradient Descent (SGD)}, \textit{Stochastic Average Gradient (SAG), Stochastic Variance Reduced Gradient (SVRG)}) to attain a relative $\theta$ accuracy per local subproblem. Then, for strongly convex objectives, the general upper bound on the number of iterations is dependent on local relative $\theta$ accuracy of the local subproblem and the global model's accuracy $\epsilon$ as \cite{konecny2016federated}:
		\begin{align}
		I^{\textrm{g}}(\epsilon,\theta) &=  \frac{\zeta \cdot\log (\frac{1}{\epsilon})}{1-\theta}, \label{relation:1}
		\end{align}	
	where the local relative accuracy measures the quality of the local solution as defined in the earlier paragraphs. Further, in this formulation, we have replaced the term $\mathcal{O}(\log (\frac{1}{\epsilon}))$ in the numerator with $\zeta \cdot\log (\frac{1}{\epsilon})$, for a constant $\zeta > 0$. For fixed iterations $I^{\textrm{g}}$ at the MEC server to solve the global problem, we observe in \eqref{relation:1} that a very high local accuracy (small $\theta$) can significantly improve the global accuracy $\epsilon$. However, each client $k$ has to spend excessive resources in terms of local iterations, $I_k^{\textrm{l}}$ to attain a small $\theta_k$ accuracy as
	\begin{align}
	I_k^{\textrm{l}}(\theta_k) = \gamma_k\log \bigg(\frac{1}{\theta_k}\bigg), \label{relation:2}
	\end{align}
	where $\gamma_k>0$ is a parameter choice of client $k$ that depends on the data size and condition number of the local subproblem \cite{konevcny2017semi}. 
	Therefore, to address this trade-off, MEC server can setup an economic interaction environment (a crowdsourcing framework) to motivate the participating clients for improving the local relative $\theta_k$ accuracy. Correspondingly, with the increased reward, the participating clients are motivated to attain better local $\theta_k$ accuracy, which as observed in (\ref{relation:1}) can improve the global $\epsilon$ accuracy for a fixed number of iterations $I^{\textrm{g}}$ of the MEC server to solve the global problem. In this scenario, to capture the statistical and system-level heterogeneity, the corresponding performance bound in (9) for heterogeneous  responses $\theta_{k}$ can be modified considering the worst-case response of the participating client as 
	\begin{equation}
	I^{\textrm{g}}(\epsilon,\theta_k) = \frac{\zeta \cdot \log(\frac{1}{\epsilon})}{1 - {\max}_k\ \theta_k}, \forall k \in \mathcal{K}.
	\label{eq:het_response}
	\end{equation}
	
	Fig. \ref{fig:message_pass} describes an interaction environment incorporating crowdsourcing framework and FL setting. In the following section, we will further discuss in details about the proposed incentive mechanism, and present the interaction between MEC server and participating clients as a two-stage Stackelberg game.

	\subsection{Cost Model}
	Training on local data for a defined accuracy level incurs a cost for the participating clients. We discuss its significance  with two typical costs: the computing cost and the communication cost.
	
	\textbf{Computing cost:} This cost is related to the number of iterations performed by client $k$ on its local data to train the local model for attaining a relative accuracy of $\theta_k$ in a single round of communication. With \eqref{relation:2}, we define the computing cost for client $k$ when it performs computation on its local data $\mathcal{D}_k$.
	
	\textbf{Communication cost:} This cost 
	is incurred when client $k$ interacts with MEC server for parameter updates to maintain  $\theta_k$ accuracy. During a round of communication with the MEC server, let $e_k$ be the size (in bits) of local parameters $\Delta \phi_{[k]}, k \in \mathcal{K}$ in a floating point representation produced by the participating client $k$ after processing a mini-batch \cite{iandola2016firecaffe}. 
	While $e_k$ is the same for all the participating clients under a specified learning setting of the global problem, each participating client $k$ can invest resources to attain specific $\theta_k$ as defined in \eqref{relation:2}. Although the best choice would be to choose $\theta_k$ such that the local solution time is comparable with the time expense in a single communication round, larger $\theta_k$ will induce more rounds of interaction between clients until global convergence, as formalized in \eqref{relation:1}. 
	
	With the inverse relation of global iteration upon local relative accuracy in \eqref{relation:1}, we can characterize the total communication expenditure as
	\begin{equation}
	T(\theta_k) =  \frac{T_k}{(1 - \theta_k)},
	\label{eq:com_exp}
	\end{equation}
	where $T_k$ as the time required for the client $k$ to communicate with MEC server in each round of model's parameter exchanges. Here, 	we normalize $\zeta >0$ in \eqref{relation:1} to 1 as the constant can be absorbed into $T_k$ for each round of model's parameter exchanges when we characterize the communication expenditure in \eqref{eq:com_exp}. Using first-order Taylor's approximation\footnote{First-order taylor's approximation for  $f(\theta) = \frac{1}{1 - \theta}$ is $ f(\theta)\mid_{\theta = a } = f(a) + f'(a)(\theta - a)$. For small $\theta$, the approximation results $f(\theta)\mid_{\theta = 0 } =  1 + \theta.$ }, we can approximate the total communication cost as $T(\theta_k) = T_k \cdot (1 + \theta_k)$. We assume that clients are allocated orthogonal sub-channels so that there is no interference between them\footnote{Note that the scenario of possible delay introduced with interference on poor wireless uplink channel can affect the local model update time. This can be mitigated by adjusting maximum waiting time as in \cite{kim2018device} at MEC.}. Therefore, the instantaneous data rate for client $k$ 
	can be expressed as 
	\begin{equation}
	R_k = B \log_2 \left(1 + \frac{p_k |G_k|^2}{\mathcal{N}_k}\right), \forall k \in \mathcal{K},
	\label{eq:instantaneous_rate}
	\end{equation}
	where $B$ is the total bandwidth allocated to the client $k$, $p_k$ is the transmission power of the client $k$, $|G_k|^2$ is the channel gain between participating client $k$ and the base station, and  $\mathcal{N}_k$ is the Gaussian noise power at client $k$. 
	Then for client $k$, using \eqref{eq:instantaneous_rate}, we can characterize $T_k$ for each round of communication with the MEC server to upload the required updates as 
	\begin{equation}
	T_k = \frac{e_k}{B \log_2 \left(1 + \frac{p_k |G_k|^2}{\mathcal{N}_k}\right)}, \forall k \in \mathcal{K}.
	\label{eq:communication_cost}
	\end{equation}
	\eqref{eq:communication_cost} provides the dependency of $T_k$ on wireless conditions and network connectivity.
	
	Assimilating the rationale behind our earlier discussions, for a participating client with evaluated $T_k$, the increase in value of $\theta_k$ (poor local accuracy) will contribute for a larger communication expenditure. This is because the participating client has to interact more frequently with the MEC server (increased number of global iterations) to update its local model parameter for attaining relative $\theta_k$ accuracy. Further, the authors in \cite{dinh2019federated} have provided the convergence analysis to justify this relationship and the communication cost model, though with a different technique.  
	
	Therefore, the participating client $k$'s cost for the relative accuracy level $\theta_k$ on the local subproblem is
	\begin{align}
	\begin{split}
	C_k(\theta_k)& = 
	(1+\theta_k ) \cdot \left(\nu_k \cdot T_k + (1 - \nu_k)\cdot \gamma_k  \log\bigg(\frac{1}{\theta_k}\bigg)\right), \\ 
	\end{split}
	\label{totalcost}
	\end{align}
	where 0 $\le\nu_k\le$1 is the normalized monetary weight for communication and computing costs (i.e., \$/ rounds of iteration).
	A \textbf{smaller value of relative accuracy $\theta_k$ indicates a high local accuracy}. Thus, there exists a trade-off between the communication and the computing cost (\ref{totalcost}). A participating client can adjust its preference on each of these costs with the weight metric $\nu_k$. The higher value of $\nu_k$ emphasizes on the larger rounds of interaction with the MEC server to adjust its local model parameters for the relative $\theta_k$ accuracy. On the other hand, the higher value of $(1- \nu_k)$ reflects the increased number of iterations at the local subproblem to achieve the relative $\theta_k$ accuracy. This will also significantly reduce the overall contribution of communication expenditure in the total cost formulation for the client. Note that the client cost over iterations could not be the same. However, to make the problem more tractable, according to (9) we consider minimizing the upper-bound of the cost instead of the actual cost, similar to approach in \cite{wang2018edge}.

	\section{Incentive Mechanism for Client's Participation in  the Decentralized Learning Framework} \label{sec:Incentive_Mechanism}

	In this section, firstly, we present our motivation to realize the concept of FL by employing a crowdsourcing framework. We next advocate an incentive mechanism required to realize this setting of decentralized learning model with our proposed solution approach. 

	\subsection{Incentive Mechanism: A Two-Stage Stackelberg Game Approach}

	The MEC server will allocate reward to the participating clients to achieve optimal local accuracy in consideration for improving communication efficiency of the system. That means, the MEC server will plan to incentivize clients for maximizing its own benefit, i.e., an improved global model. Consequently, upon receiving the announced reward, any rational client will individually maximize their own profit. Such interaction scenario can be realized with a Stackelberg game approach. 
	
	Specifically, we formulate our problem as a two-stage Stackelberg game between the MEC server (leader) and participating clients (followers). Under the crowdsourcing framework, the MEC server designs an incentive mechanism for participating clients to attain a local consensus accuracy level\footnote{It signifies the agreement among the participating clients on the quality of solution at the local subproblems for building a high-quality centralized learning model.} on the local models while improving the performance of a centralized model. The MEC server cannot directly control the participating clients to maintain a local consensus accuracy level, and requires an effective incentive plan to enroll clients for this setting. 
	
	\textbf{Clients (Stage II):} The MEC server has an advantage, being a leader with the first-move advantage influencing the followers for participation with a local consensus accuracy. It will at first announce a uniform reward rate\footnote{Prominently, two kinds of pricing scheme exist at present following different design goals: uniform pricing and discriminatory or differentiated pricing \cite{liu2017interference}. The differentiated pricing scheme is more efficient, but also requires more information and higher complexity than the uniform pricing \cite{li2016pricing}, \cite{faltings2014incentive}. Therefore, based upon offered motivations and benefits, our proposed crowdsourcing framework follows a platform-centric model to train a high quality global model with low complexity, less information exchange by using the uniform pricing scheme.} (e.g., a fair data package discount as \$/accuracy level) $r>0$ for the participating clients. Given $r$, at Stage II, a rational client $k$ will try to improve the local model's accuracy for maximizing its net utility by training over the local data with global parameters. The proposed utility framework incorporates the cost involved while a client tries to maximize its own individual utility.   
	
	\textit{Client Utility Model:}
	We use a valuation function  $v_k(\theta_k)$ to denote the model's effectiveness that explains the valuation of the client $k $ when relative  $ \theta_k$  accuracy is attained for the local subproblem.
	\\
	\textbf{Assumption 1.} The valuation function $v_k(\theta_k)$ is a linear, decreasing function with $\theta_k>0$, i.e., $v_k(\theta_k) = (1- \theta_{k})$. Intuitively, for a smaller relative accuracy at the local subproblem, there will be an increase in the reward for the participating clients. 
			
	Given $r > 0$, each participating client $k$'s strategy is to maximize its own utility as follows: 
	\begin{align}
	\begin{split}
	\underset{ 0 \le \theta_k \le  1 }{\text{max}} \qquad &
	u_k(r,\theta_k) = r (1 - \theta_k) - C_k(\theta_k),
	\end{split}
	\label{client_utility_max}
	\end{align}
	given cost $C_k(\theta_k)$ as \eqref{totalcost}. The feasible solution is always restricted to the value less than 1 (i.e., without loss of generality, for $ \theta_k > 1$, it violates the participation assumption for the crowdsourcing framework). Therefore, problem (\ref{client_utility_max}) can be represented as 
	\begin{align}
	\begin{split}
	\underset{\theta_k > 0}{\text{max}} \qquad &
	u_k(r,\theta_k) = r(1-\theta_k) - C_k(\theta_k),\forall k \in \mathcal{K}.
	\end{split}
	\label{client_utility_max_relaxed}
	\end{align}
\color{black}
	Also, we have $C^{''}_k(\theta_k) > 0$, which means $C_k(\theta_k)$ is a strictly convex function. Thus, there exists a unique solution $\theta_k^*(r), \forall k.$
	
	\textbf{MEC Server(Stage I):} Knowing the response (strategy) of the participating clients, the MEC can evaluate an optimal reward rate $r^*$ to maximize its utility. The utility $U(\cdot)$ of MEC server can be defined in relation to the satisfaction measure achieved with local consensus accuracy level. 
	
	\textit{MEC Server Utility Model:} We define $x(\epsilon)$ as the number of iterations required for an arbitrary algorithm to converge to some $\epsilon$ accuracy. We similarly define $I^{\textrm{g}}(\epsilon, \theta)$ as global iterations of the framework to reach a relative $\theta$ accuracy on the local subproblems. 
	
	From this perspective, we require an appropriate utility function $U(\cdot)$ as the satisfaction measure of the framework with respect to the number of iterations for achieving $\epsilon$ accuracy. In this regard, use the definition of the number of iterations for $\epsilon$ accuracy as
	$$x(\epsilon) = \zeta \cdot\log\bigg(\frac{1}{\epsilon}\bigg).$$
	Due to large values of iterations, we approximate $x(\epsilon)$ as a continuous value, and with the aforementioned relation, we choose $U(\cdot)$ as a strictly concave function of $x(\epsilon)$ for $\epsilon \in [0,1]$, i.e., with the increase in $x(\epsilon)$, $U(\cdot)$ also increases. Thus, we propose $U(x(\epsilon))$ as the normalized utility function bounded within $[0,1]$ as 
	\begin{equation}
	U(x(\epsilon)) = 1 - 10^{-(ax(\epsilon)+b)},\ \ \ a \ge 0,b \le 0,
	\label{MEC_Utility}
	\end{equation}
	which is strictly increasing with $x(\epsilon)$, and represents the satisfaction of MEC increase with respect to accuracy $\epsilon$.

	\begin{figure}[t!]
		\centering
		\includegraphics[width=3in]{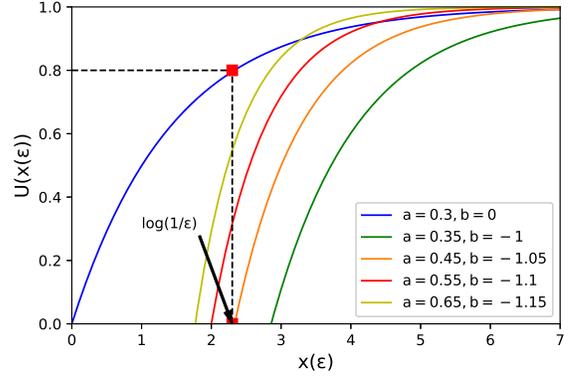}
		\caption{MEC utility $U(\cdot)$ as a function of $\epsilon$ with different parameter values of $a,b$.}
		\label{fig_utility}
	\end{figure}
	
	As for the global model, there exists an acceptable value of threshold accuracy measure correspondingly reflected by $x_{\min}(\epsilon)$. This suggests the possibility of near-zero utility for MEC server for failing to attain such value.
	
	Fig. \ref{fig_utility} depicts our proposed utility function, a concave function of $x(\epsilon)$ with parameters $a$ and $b$ that reflect the required behavior of the utility function defined in (\ref{MEC_Utility}). In Fig. \ref{fig_utility}, we can observe that larger value of $a$ means smaller iterations requirement and larger values of $b$ introduces flat curves suggesting more flexibility in accuracy. So we can analyze the impact of parameters $a$ and $b$ in (\ref{MEC_Utility}), and set them to model the utility function for the MEC server as per the design requirements of the learning framework. Furthermore, in our setting, $I^{\textrm{g}}(\epsilon,\theta)$ can be elaborated with a upper bound (maximum global iterations, $\delta$) as
	\begin{align}
	& I^{\textrm{g}}(\epsilon,\theta) = \frac{x(\epsilon)}{1-\theta} \le \delta.
	\label{CoCoA_bounded}		
	\end{align}
	(\ref{CoCoA_bounded}) explains the efficiency paradigm of the proposed framework in terms of time required for the convergence to some accuracy $\epsilon$. If $\tau^{\textcolor{blue}{\textrm{l}}}(\theta)$ is the time per iteration to reach a relative $\theta$ accuracy at  a local subproblem and $T(\theta)$ is the communication time required during a single iteration for any arbitrary algorithm, then  we can analyze the result in (\ref{CoCoA_bounded}) with the efficiency of the global model as
	\begin{align}
	& I^{\textrm{g}}(\epsilon, \theta) \cdot (T(\theta)+ \tau^{\textrm{l}}(\theta)).
	\label{efficieny}
	\end{align}
	Because the cost of communication is proportional to the speed and energy consumption in a distributed scenario \cite{bao2018online}, the bound defined in ($\ref{CoCoA_bounded}$) explains the efficiency in terms of MEC server's resource restriction for attaining $\epsilon$ accuracy. In this regard, the corresponding analysis of (20) is presented in the upcoming sub-section with several case studies.
	
	The utility of the MEC server can therefore be defined for the set of measured best responses $\boldsymbol{\theta}^*$ as 
	$$\mathcal{U}(x(\epsilon),r| \boldsymbol{\theta}^*) = \beta \left(1 - 10^{-(ax(\epsilon)+b)} \right)  - r \sum_{k \in \mathcal{K}} (1 - \theta^*_k(r)),$$
	where $\beta > 0$ is the system parameter \footnote{Note that $\beta > 0$ characterizes a linear scaling metric to the utility function which can be set arbitrarily and will not alter our evaluation. Equivalently, it can be understood as the MEC server's physical resource consignments for the FL that reflects the satisfaction measure of the framework.}, and $r \sum_{k \in \mathcal{K}} (1 - \theta^*_k(r))$ is the cost spent for incentivizing participating clients in the crowdsourcing framework for FL.
	So, for the measured $\boldsymbol{\theta}^*$ from the participating clients at MEC server, the utility  maximization problem can be formulated as follows:
	\begin{align}
	\begin{split}
	\underset{r \ge 0, x(\epsilon) }{\text{max}} \qquad &
	\mathcal{U}(x(\epsilon),r| \boldsymbol{\theta}^*),\label{eq:MEC_Utility}
	\end{split}	
	\\
	\text{s.t.}\qquad &  \frac{x(\epsilon)}{1-{\max}_k\ \theta^*_k(r)} \le \delta. \label{eq : constraint}
	\end{align}\color{black}
	In constraint \eqref{eq : constraint}, ${\max}_k\ \theta^*_k(r)$ characterizes the worst case response for the server side utility maximization problem with the bound on permissible global iterations. Note that MEC adapts admission control strategy (discussed in Section VII) to improve the number of participation for maximizing its utility. In fact, MEC has to increase the reward rate to maintain a minimum number of participation (at least two) to realize the distributed optimization setting in FL. In addition to this, the framework may suffer from slower convergence due to fewer participation. Thus, MEC will avoid deliberately dropping the clients to achieve a faster consensus with \eqref{eq : constraint}.	 
	
	Furthermore, using the relationship defined in (\ref{CoCoA_bounded}) between $x(\epsilon)$ and relative $\theta$ accuracy for the subproblem, we can analyze the impact of responses $\boldsymbol{\theta}$ on MEC server's utility in a FL setting with the constraint \eqref{eq:het_response}. To be more specific about this relation, we can observe that with the increased value of $(1- \theta)$, i.e., lower relative accuracy (high local accuracy), the MEC server can attain better utility due to corresponding increment in value of $x(\epsilon)$. Note that in the client cost problem, $x(\epsilon)$ is treated as a constant provided by the MEC problem, and can be ignored for solving \eqref{client_utility_max}. 

	\newtheorem{mylem}{Lemma}
	\begin{mylem}
		The optimal solution $x^*(\epsilon)$ for (\ref{eq:MEC_Utility}) can be derived as $\delta (1- {\max}_k \ \theta^*_k(r))$. 
	\end{mylem}

	\begin{IEEEproof}
		See Appendix A.
	\end{IEEEproof}
	Therefore, for the given $\boldsymbol{\theta}^*(r)$, we can formalize (\ref{eq:MEC_Utility}) as  
	\begin{align}
	\begin{split}
	\underset{r \ge 0}{\text{max}} \qquad & \beta \left(1 - 10^{-(ax^*(\epsilon)+b)} \right)  - r \sum_{k \in \mathcal{K}} (1 - \theta^*_k(r)). \label{MEC_Opt} 
	\end{split}	
	\end{align}\color{black}
	\textbf{Stackelberg Equilibrium.} With a solution to MEC server's utility maximization problem, $r^*$ we have the following definition.
	
	\textbf{Definition 1.} For any values of $r$, and $\boldsymbol{\theta}$, $(r^*, \boldsymbol{\theta}^*)$ is a Stackelberg equilibrium if it satisfies the following conditions:
	\begin{align}
	&\mathcal{U}(r^*,\boldsymbol{\theta}^*) \ge \mathcal{U}(r,\boldsymbol{\theta}^*), \label{relation_1}\\
	& u_k(\theta_k^*,r^*) \ge u_k(\theta_k,r^*), \ \forall k.  \label{relation_2}
	\end{align}
	
	Next, we employ the backward-induction method to analyze the Stackelberg equilibria: the Stage-II problem is solved at first to obtain $\boldsymbol{\theta}^*$, which is then used for solving the Stage-I problem to obtain $r^*$.

	\subsection{Stackelberg Equilibrium: Algorithm and Solution Approach}

	Intuitively, from (\ref{CoCoA_bounded}), we see that the server can evaluate the maximum value of $x(\epsilon)$ required for attaining accuracy $\epsilon$ for the centralized model while maintaining relative accuracy $\theta_{\textrm{th}}$ amongst the participating clients. Here, $\theta_{\textrm{th}}$ is a consensus on a maximum local accuracy level amongst participating clients, i.e., the local subproblems will maintain at least $\theta_{\textrm{th}}$ relative accuracy. So, with the measured responses $\boldsymbol{\theta}$ from the participating clients, the server can design a proper incentive plan to improve the global model while maintaining the worst case relative accuracy ${\max}_k \ \theta^*_k$ as $\theta_{\textrm{th}}$ for the local model. 
	
	Since the threshold accuracy $\theta_{\textrm{th}}$ can be adjusted by the MEC server for each round of solution, each participating client will maintain a response towards the maximum local consensus accuracy $\theta_{\textrm{th}}$. This formalizes the client's selection criteria [see Remark 1.] which is sufficient enough for the MEC server to maintain the accuracy $\epsilon$. We also have the lower bound related with the value of $x_{\min}(\epsilon)$ for equivalent accuracy $\epsilon_{\max}$ while dealing with the client's responses $\boldsymbol{\theta}$, i.e.,
		\begin{align}
		\begin{split}
		\log\left( \frac{1}{\epsilon_{\max}}\right) \le \frac{x(\epsilon)}{(1-\theta_{\textrm{th}})} \le \delta_{\max}. 
		\label{CoCoA_detailed}
		\end{split}
		\end{align}
		where $\delta_{\max}$ is the maximum permissible upper bound to the global iterations.
	
	As explained before and with (\ref{CoCoA_detailed}), the value of $\theta_{\textrm{th}}$ can be varied (lowered) by MEC server to improve the overall performance of the system. For a worst case scenario, where the offered reward $r$ for the client $k$ is insufficient to motivate it for participation with improved local relative accuracy, we might have ${\max}_k\ \theta^*_k(r) = 1$, i.e., $\theta_{\textrm{th}} = 1$, no participation.

	\begin{mylem}
		For a given reward rate $r$, and $T_k$ which is determined based upon the channel conditions \eqref{eq:communication_cost}, we have the unique solution $\theta^*_k(r)$ for the participating client satisfying following relation:
		\begin{equation}
		\begin{aligned}
		g_k(r) = \log(e^{1/\theta^*_k (r)}\theta^*_k (r)), \forall k \in \mathcal{K},
		\label{eq:client_response}
		\end{aligned}
		\end{equation}
		for $g_k(r) \geq 1$, where, 
		$$g_k(r) = \left[ \frac{r + \nu_k T_k}{(1 - \nu_k)\gamma_k} -1 \right].$$
		\label{lemma2}	
	\end{mylem}

	\textit{
		Proof:} Because $C_k^{''}(\theta_k) > 0$ for $\theta_k > 0$, (\ref{client_utility_max_relaxed}) is a strictly convex function resulted as a linear plus convex structure. Therefore, by the first-order condition, (\ref{client_utility_max_relaxed}) can be deduced as 
	\begin{align}
	\begin{split}
	\frac{\partial u_k(r,\theta_k)}{\partial \theta_k}&  = 0 \\ 
	& \Leftrightarrow  \frac{1}{\theta_k} - \log\bigg(\frac{1}{\theta_k}\bigg) = \left[\frac{r + \nu_kT_k}{(1- \nu_k)\gamma_k} -1\right],\\
	& \Leftrightarrow \log(e^{1/\theta_k}\theta_k) = g_k(r).
	\label{Lemma:proof}
	\end{split}
	\end{align}
	We observe that Lemma \ref{lemma2} is a direct consequence of the solution structure derived in \eqref{Lemma:proof}. Hence, we conclude the proof.
	
	From Lemma \ref{lemma2}, we have some observations with the definition of $g_k(r)$ for the response of the participating clients. First, we can show that $\theta^*_k$ is larger for the poor channel condition on a given reward rate. Second, in such scenario, with the increase in reward rate, say for $g_k(r) > 2$ the participating clients will iterate more during their computation phase resulting in lower $\theta^*_k$. This will reduce the number of global iterations to attain an accuracy level for the global problem.
	
	We can therefore characterize the participating client $k$'s best response under the proposed framework as 
		\begin{equation}
		\begin{aligned}
		\theta_k^*(r) & = \min \left\lbrace \hat{\theta}_k (r) \mid_{g_k(r) = \log(e^{1/\hat{\theta}_k(r)}\hat{\theta}_k(r))}, \theta_{\textrm{th}} \right\rbrace, \forall k.
		\label{client_best_response}
		\end{aligned}
		\end{equation}		
		(\ref{client_best_response}) represents the best response strategy for the participating client $k$ under our proposed framework. Intuitively, exploring the logarithmic structure in \eqref{eq:client_response}, we observe that the increase in incentive $r$ will motivate participating clients to increase their efforts for local iteration in one global iteration. This is reflected by a better response, i.e., a lower relative accuracy (high local accuracy) during each round of communication with the MEC server.	
	\begin{figure}[t!]
		\centering
		\includegraphics[width=3in]{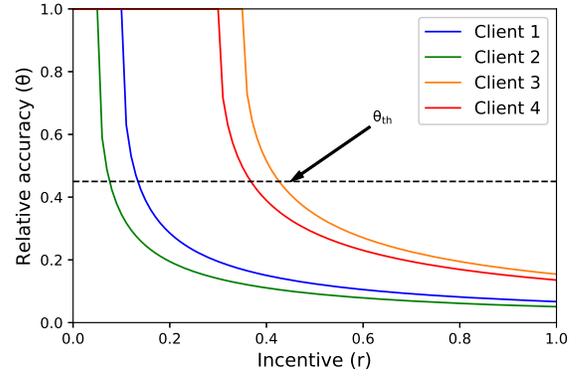}
		\caption{An illustration showing participating clients response over the offered reward rate.}
		\label{fig: Clients_Response}
	\end{figure}
	
	Fig. \ref{fig: Clients_Response} illustrates such strategic responses of the participating clients over an offered reward for a given configuration. In this scenario, to elaborate the best response strategy as characterized in \eqref{client_best_response}, we have considered four participating clients with different preferences (e.g., Client 3 being the most reluctant participant). We observe that Client 3 seeks more incentive $r$ to maintain comparable accuracy level as Client 1. Further, we consider the trade off between communication cost and the computation cost as discussed with the relation in \eqref{totalcost}. These costs are complementary in relation by $\nu_k$, and for each client $k$ their preferences upon these costs are also different. For instance, the higher value of $\nu_k$ for client $k$ emphasizes on the increased number of communication with the MEC server to improve the local relative accuracy $\theta_k$.
	
	\begin{figure*}[t!]
		\centering
		\begin{subfigure}[b]{0.3\textwidth}
			\centering
			\includegraphics[width=2.4in]{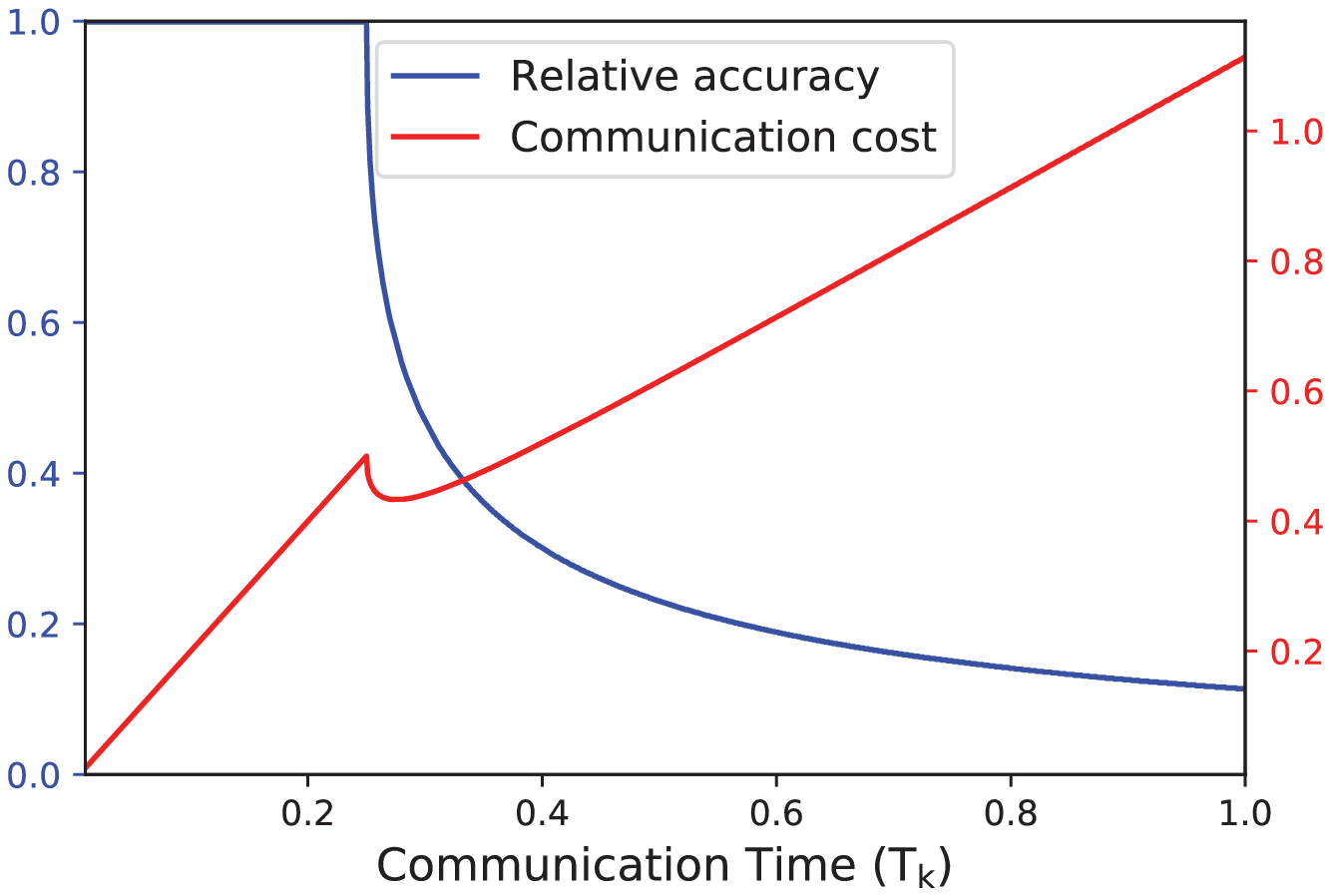}
			\caption{}
		\end{subfigure}%
		\hfill
		\begin{subfigure}[b]{0.3\textwidth}
			\centering
			\includegraphics[width=2.4in]{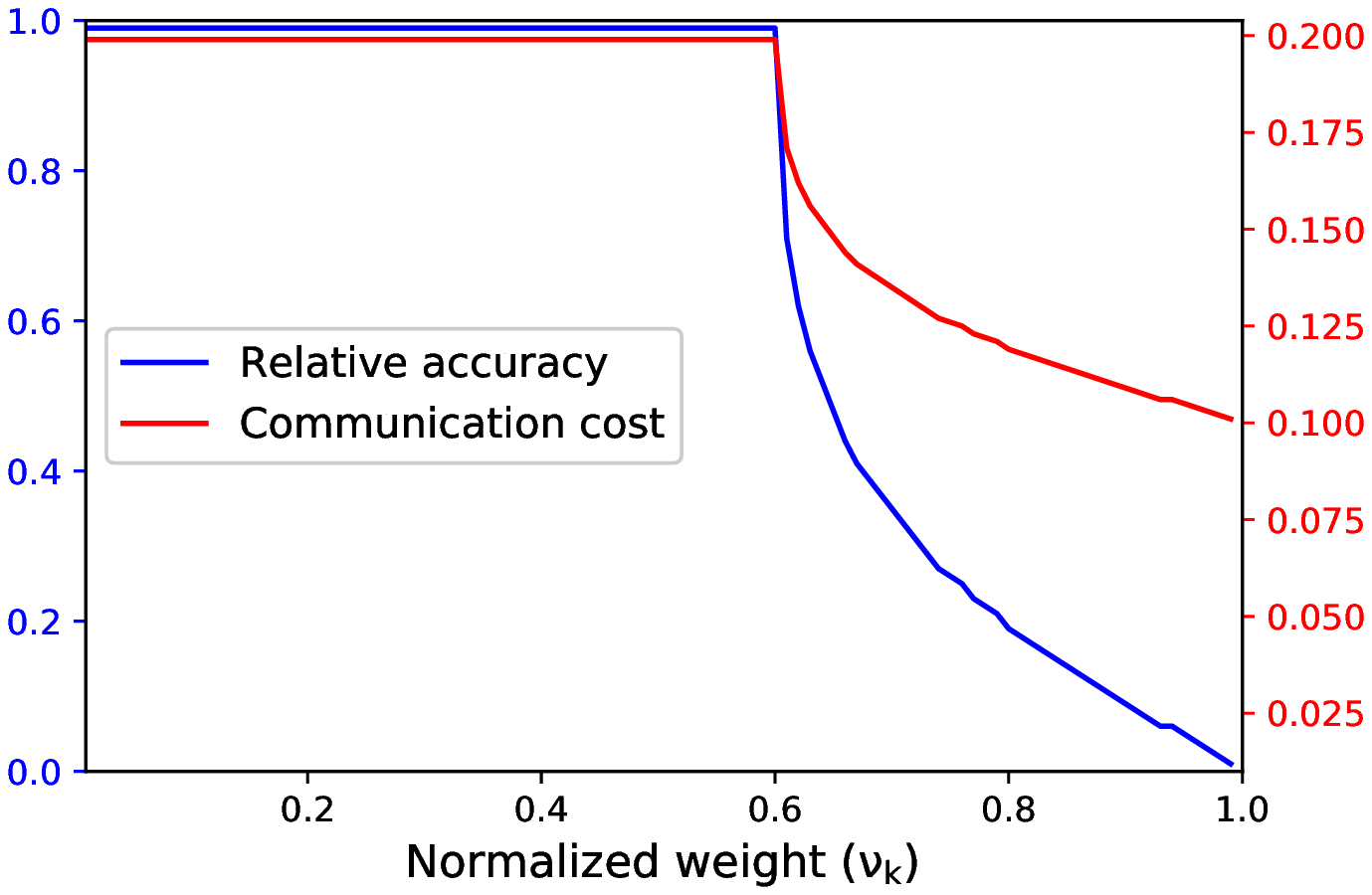}
			\caption{}
			
		\end{subfigure}
		\hfill
		\begin{subfigure}[b]{0.3\textwidth}
			\centering
			\includegraphics[width=2.4in]{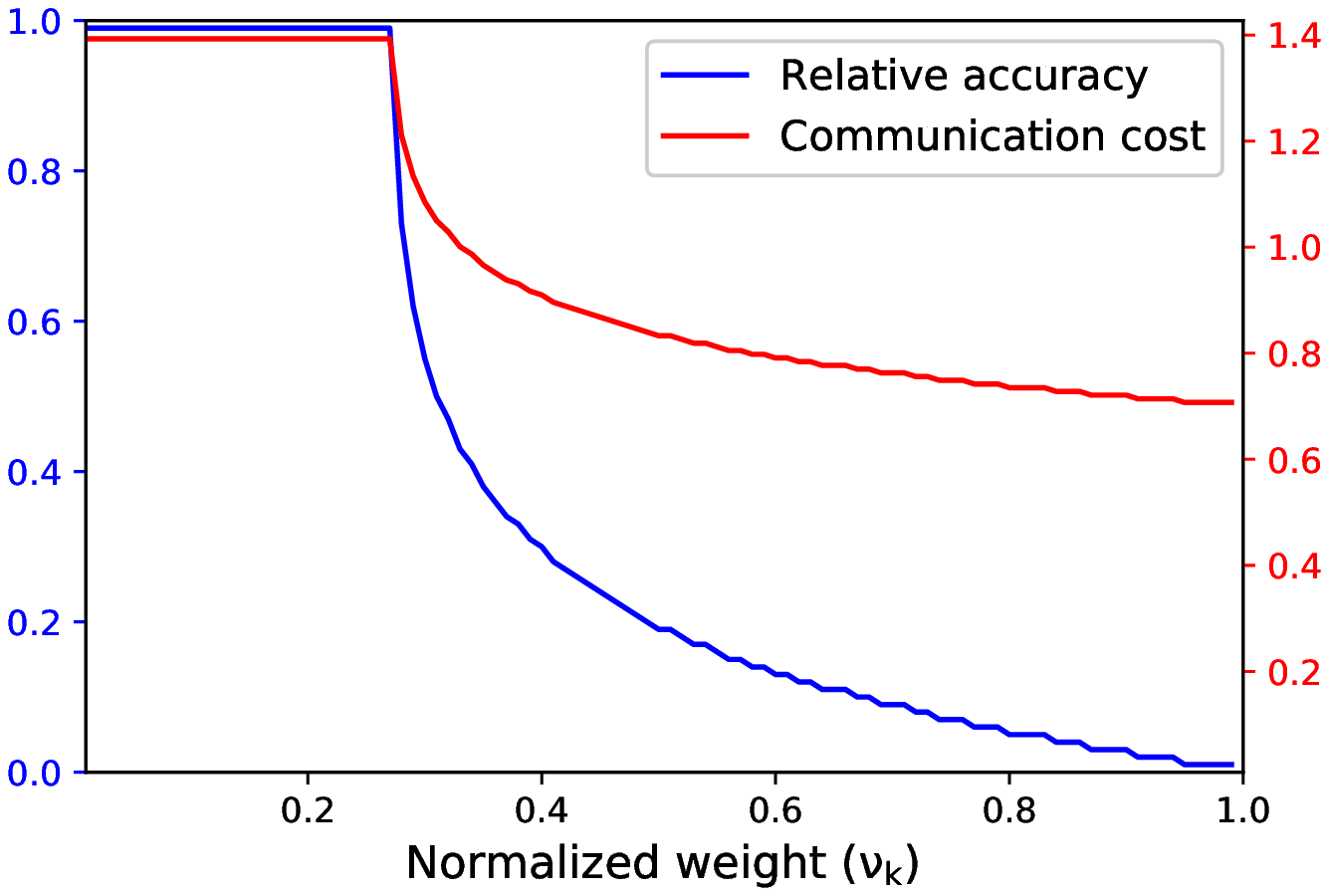}
			\caption{}
		\end{subfigure}
		\centering
		\caption{Solution Analysis \eqref{eq:client_response} (Left Y-axis: Relative accuracy, Right Y-axis: Communication cost): (a) impact of communication adversity on local relative accuracy for a constant reward (b) normalized weight versus relative accuracy for a fair data rate (quality communication channel) (c) normalized weight versus relative accuracy for an expensive data rate.} 
		\label{fig:response_scenarios}
	\end{figure*}
	\begin{figure*}[t!]
		\centering
		\begin{subfigure}[b]{0.3\textwidth}
			\centering
			\includegraphics[width=2.4in]{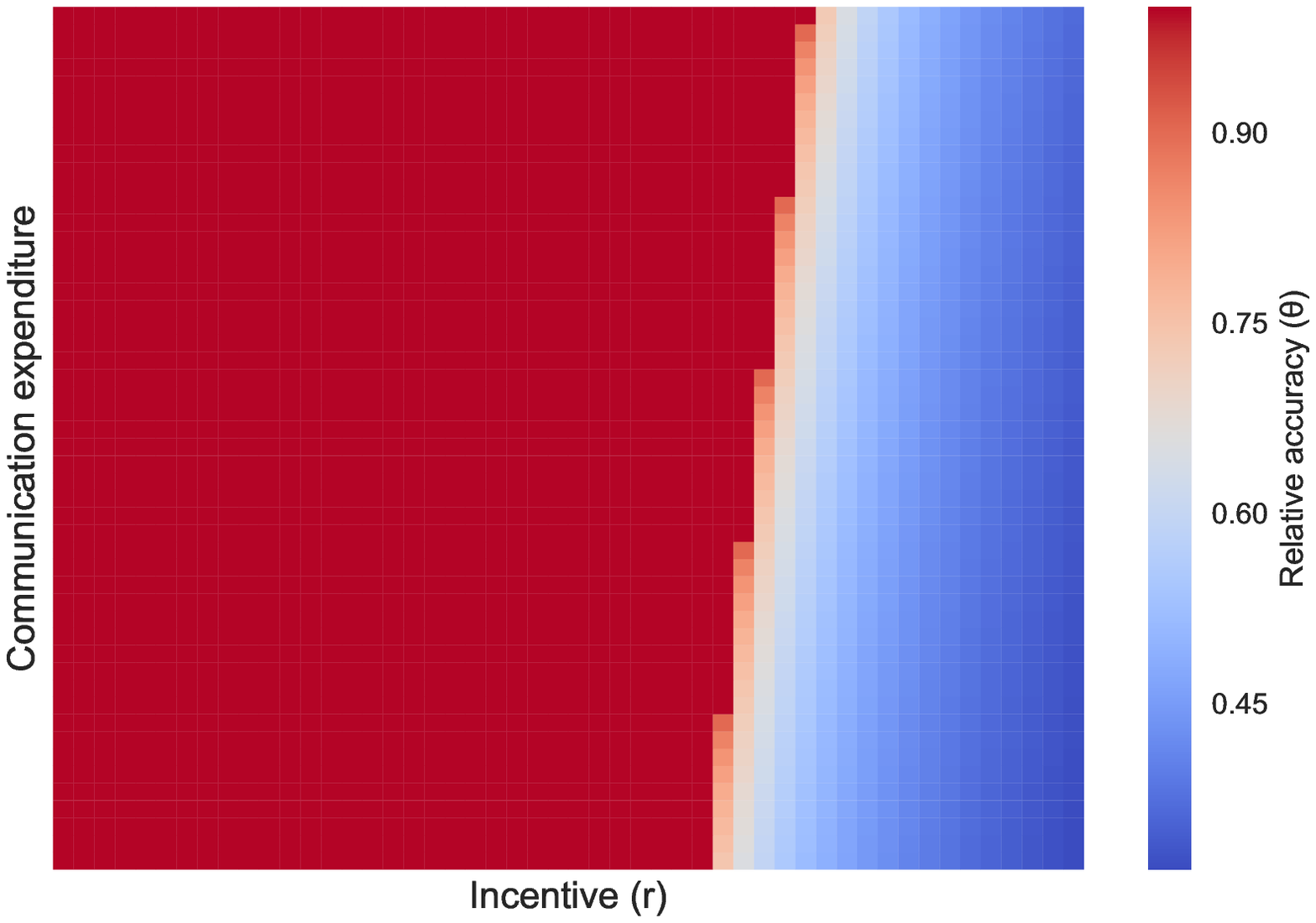}
			\caption{}
		\end{subfigure}%
		\hfill
		\begin{subfigure}[b]{0.3\textwidth}
			\centering
			\includegraphics[width=2.4in]{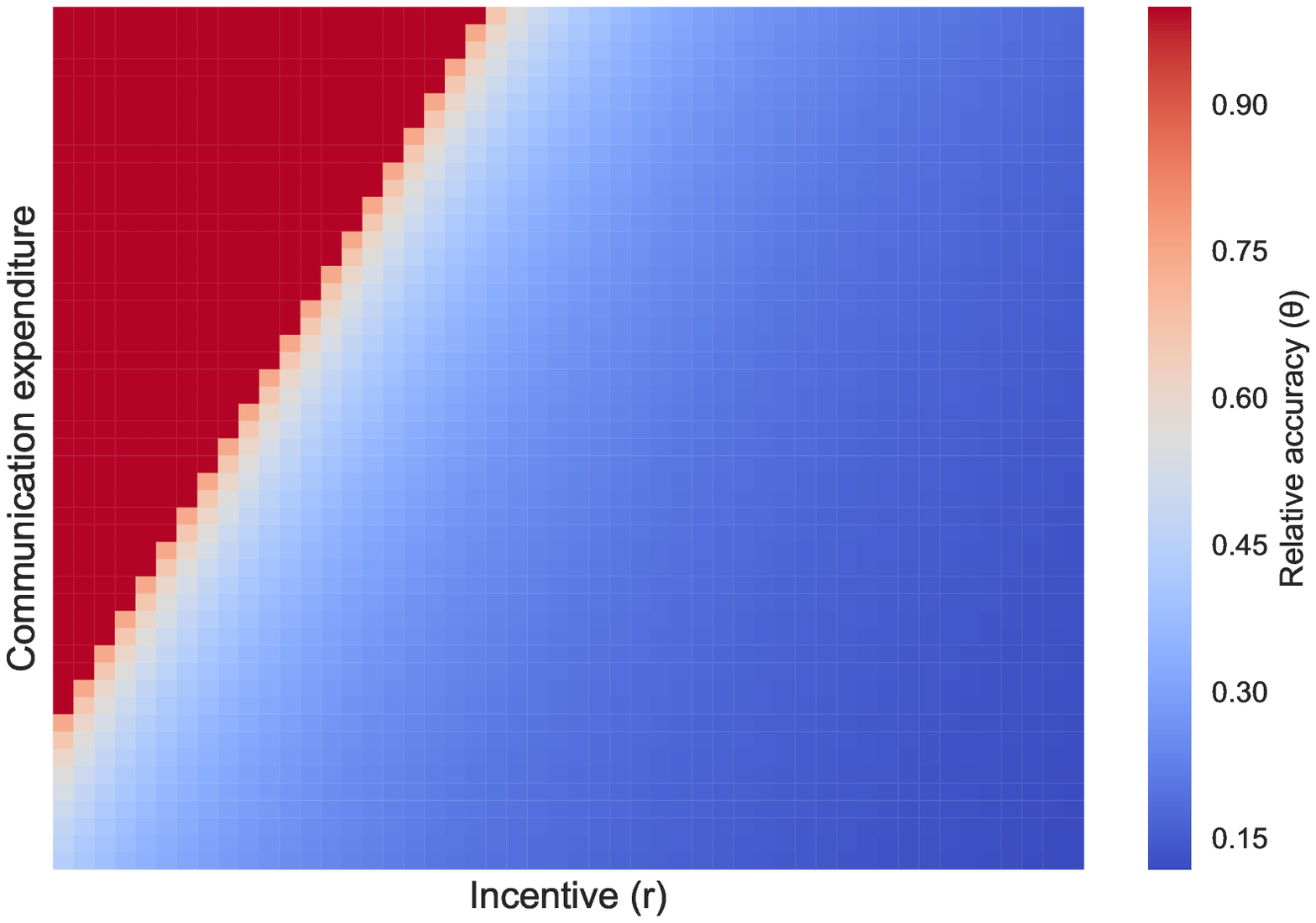}
			\caption{}	
		\end{subfigure}
		\hfill
		\begin{subfigure}[b]{0.3\textwidth}
			\centering
			\includegraphics[width=2.4in]{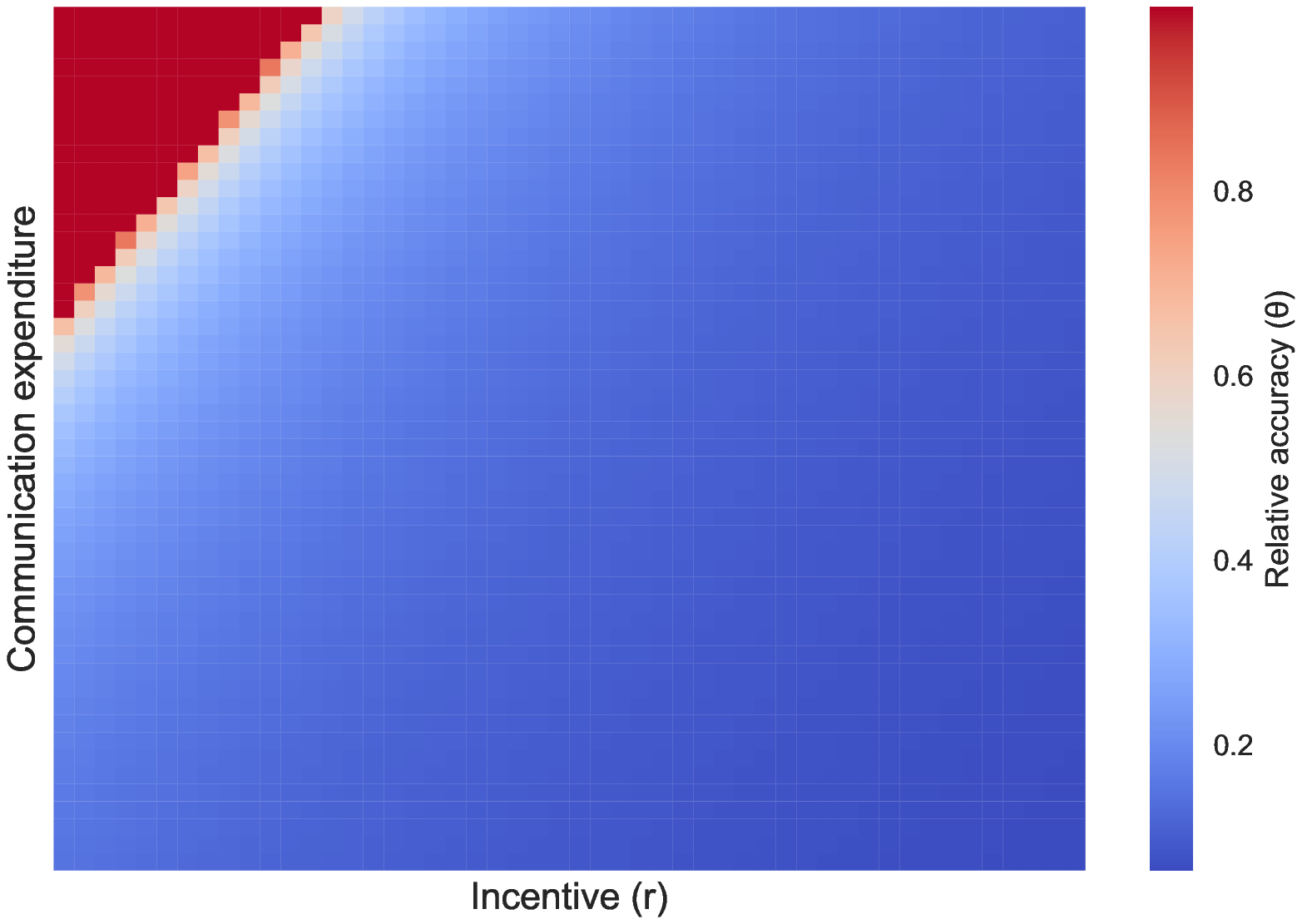}
			\caption{}
		\end{subfigure}
		\centering
		\caption{Case Study: impact of communication cost and offered reward rate $r$ for different values of normalized weight (preferences), $\nu_k$ defining client's categories (a) Reluctant, $\nu_k = 0.1$ (b) Rational, $\nu_k =0.5$ (c) Sensitive, $\nu_k =0.7$. X-axis shows the increase in incentive (r) value from left-to-right, and the y-axis defines the increase in value of communication expenditure (top-to-bottom).}
		\label{fig:response_case_study}
	\end{figure*}
	In Fig. \ref{fig:response_scenarios}, we briefly present the solution analysis to \eqref{eq:client_response} with the impact of channel condition (we define it as communication adversity) on the local relative accuracy for a constant reward. For this, in Fig. \ref{fig:response_scenarios}a we consider a participating client with the fixed offered reward setting $r$ from uniformly distributed values of 0.1 to 5. We use normalized $T_k$ parameter for a client $k$ to illustrate the response analysis scenario. In Fig. \ref{fig:response_scenarios}b and  Fig. \ref{fig:response_scenarios}c, $T_k$ is uniformly distributed on [0.1, 1], and $\nu_k$ is set at 0.6. Intuitively, as in  Fig. \ref{fig:response_scenarios}a, the increase in communication time $T_k$ for a fixed reward $r$ will influence participating clients to iterate more locally for improving local accuracy than to rely upon the global model, which will minimize their total cost. Under this scenario, we observe the increase in communication cost with the increase in communication time $T_k$. Thus, the clients will iterate more locally. However, the trend is significantly affected by normalized weights $\nu_k$, as observed in  Fig. \ref{fig:response_scenarios}b and  Fig. \ref{fig:response_scenarios}c. For a larger value of $T_k$ (poor channel condition) as in the case of  Fig. \ref{fig:response_scenarios}c, increasing the value of $\nu_k$, i.e., clients with more preference on the communication cost in the total cost model results to higher local iterations for solving local subproblems, as reflected by the better local accuracy, unlike in  Fig. \ref{fig:response_scenarios}b. In both cases we observe the decrease in communication cost upon participation. However, in  Fig. \ref{fig:response_scenarios}c the communication cost is higher because of an expensive data rate. Therefore, for a given $r$, client $k$ can adjust its weight metrics accordingly to improve the response $\theta_k$. 
	
	In Fig. \ref{fig:response_case_study}, we explore such behaviors of the participating clients through the heatmap plot. To explain better, we define three categories of participating clients based upon the value of normalized weights $\nu_k, \forall k$, which are their individual preferences upon the computation cost and the communication cost for the convergence of the learning framework. (i) \textit{Reluctant} clients with a lower $\nu_k$ consume more reward to improve local accuracy, even though the value of $T_k$ is larger (expensive), as observed in Fig. \ref{fig:response_case_study}a. (ii) \textit{Sensitive} clients are more susceptible towards the channel quality with larger $\nu_k$, and iterates more locally within a round of communication to the MEC server for improving local accuracy, as observed in Fig. \ref{fig:response_case_study}c. (iii) \textit{Rational} clients, as referred in Fig. \ref{fig:response_case_study}b tend to balance these extreme preferences (say $\nu_k = 0.5$ for client $k$), which in fact would be unrealistic to expect all the time due to heterogeneity in participating client's resources.
	
	To solve (\ref{MEC_Opt}) efficiently, with (\ref{client_best_response}) $\theta_k^*(r) =\min \left\lbrace \ \hat{\theta}_k (r) \mid_{g_k(r) = \log(e^{1/\hat{\theta}_k(r)}\hat{\theta}_k(r))}, \theta_{\textrm{th}} \right\rbrace, \forall k, $ we introduce a new variable $z_k$ in relation with consensus on local relative accuracy $\theta_{\textrm{th}}$,
	\begin{equation}
	z_k = \begin{cases}
	1, & \text{if } \ r > \  \hat{ r }_k ; \\
	0, & \text{otherwise}, 
	\end{cases}
	\label{z}
	\end{equation}	
	where
	$$\hat{ r }_k = \left[ g_k^{-1}(\log(e^{1/\theta_{\textrm{th}}}\theta_{\textrm{th}})) \right] $$
	is the minimum incentive value required obtained from (\ref{client_best_response}) to attain the local consensus accuracy $\theta_{th}$ at client $k$ for the defined parameters $\nu_k \ \text{and} \ T_k$. 
	
	This means, $\theta_k(r) < \theta_{\textrm{th}}$ when $z_k = 1$, and  $\theta_{\textrm{th}} \le \theta_k(r) < 1$ when $z_k=0$. MEC server can use this setting to drop the participants with poor accuracy. As discussed before, for the worst case scenario we consider $\theta_{\textrm{th}} = 1$.
	
	Therefore, the utility maximization problem can be equivalently written as  
	\begin{align}
	\begin{split}
	\underset{r,\{z_k\}_{k \in \mathcal{K}} }{\max} \qquad &
	\beta \left(1 - 10^{-(ax^*(\epsilon)+b)} \right) - r \sum_{k \in K} z_k\cdot (1 - \theta^*_k(r)),
	\end{split}\label{eq1}\\
	\text{s.t.} & \qquad r \ge 0, \label{eq2}\\
	& \qquad z_k \in \{0,1\}, \forall k. \label{eq3} 
	\end{align}
	\begin{algorithm}[t!]
		\caption{MEC Server's Utility Maximization}
		\begin{algorithmic}[1]
			\State Sort clients as with $\hat{ r }_1 < \hat{ r }_2 < \ldots < \hat{ r }_K$
			\State $\mathcal{R =\{\}}, \mathcal{A}=\mathcal{K}, j = K$
			\While{$j>0$}
			\State Obtain the solutions $r_j$ to the following problem:
			$$\underset{r \ge \hat{r}_1}{\max}\ \beta \left(1 - 10^{-(ax^*(\epsilon)+b)} \right)  - r \sum \nolimits_{k \in \mathcal{A}} (1 - \theta^*_k(r))$$ 
			\If {$r_j > \hat{ r }_j,$}{$ \ \ \mathcal{R} = \mathcal{R} \cup \{r_j\}; $}
			\EndIf
			\State $\mathcal{A} = \mathcal{A}\backslash j;$
			\State $j = j - 1;$
			\EndWhile
			\State Return $r_j \in \mathcal{R}$ with highest optimal values in problem (4).			
		\end{algorithmic}
		\label{Algorithm2}
	\end{algorithm} 
	The problem (\ref{eq1}) is a mixed-boolean programming, which may require exponential-complexity effort (i.e., $2^K$ configuration of $\{z_k\}_{k \in \mathcal{K}}$) to solve by the exhaustive search. To solve this problem with linear complexity, we refer to the solution approach as in Algorithm \ref{Algorithm2}.
	
	The utility maximization problem at MEC server can be reformulated as a constraint optimization problem (\ref{eq4}-\ref{eq5}) assuming a fixed configuration of $\{z_k =1\}_{k \in \mathcal{K}}$ as 
	\begin{align}
	\begin{split}
	\underset{r \ge 0}{\text{max}} \qquad & \beta \left(1 - 10^{-(ax^*(\epsilon)+b)} \right),
	\label{eq4}	
	\end{split}
	\\[2ex]
	\text{s.t.}\qquad & r \sum_{k \in K} (1 - \theta^*_k(r)) \le B,
	\label{eq5}
	\end{align}
	where (\ref{eq5}) is budget constraint for the problem. The second-order derivative of function $r(1 - \theta^*_k(r))$ in (\ref{eq5}) is $\frac{2\gamma_k(1- \nu_k)\nu_k T_k}{(r + \nu_k T_k)^3} > 0$, i.e., the problem (\ref{eq4}) is a convex problem and can be solved similarly with Algorithm \ref{Algorithm2} (line 4 -5).\\
	\\
	\textbf{Proposition 1.} \textit{Algorithm 2 can solve the Stage-I equivalent problem (\ref{MEC_Opt}) with linear complexity.}
	
	\textit{Proof:} As the clients are sorted in the order of increasing $\hat{r}_k$ (line 1), for the sufficient condition $r > \hat{r}_k$ resulting $z_k = 1$, the MEC's utility maximization problem reduces to a single-variable problem that can be solved using popular numerical methods.\\
	\\
	\textbf{Remark 1.} \textit{Algorithm 2 can maintain consensus accuracy by formalizing the clients selection criteria. This is because from (\ref{z}), $z_k = 1$ for $\theta_k(r) < \theta_{\textrm{th}}$, and $z_k=0$ for $\theta_{\textrm{th}} \le \theta_k(r) < 1$. Thus, MEC server uses this setting to drop the participants with $\theta_k(r)  > \theta_k^*(r) = \theta_{\textrm{th}}$.}\\
	\\	
	\textbf{Theorem 1.} \textit{The Stackelberg equilibria of the crowdsourcing framework are the set of pairs $\{r^*,\boldsymbol{\theta}^*\}$.}

	\textit{Proof:} For any given $\boldsymbol{\theta}$, it is obvious that $\mathcal{U}(r^*, \boldsymbol{\theta}) \ge \mathcal{U}(r, \boldsymbol{\theta}), \forall r$ since $r^*$ is the solution to the Stage-I problem. Thus, we have $\mathcal{U}(r^*,\boldsymbol{\theta}^*) \ge \mathcal{U}(r, \boldsymbol{\theta}^*)$. In the similar way, for any given value of $r$ and $\forall k$, we have $ u_k(r,\theta_k^*) \ge  u_k(r,\theta_k), \forall \theta_k$. Hence,  $ u_k(r^*,\theta_k^*) \ge  u_k(r^*,\theta_k)$. Combining these facts, we conclude the proof being based upon the definitions of \eqref{relation_1} and \eqref{relation_2}. 

	\section{Simulation Results and Analysis}\label{sim:Simulation_a}
	In this section, we present numerical simulations to illustrate our results. We consider the learning setting for a strongly convex model such as logistic regression, as discussed in Section III, to characterize and demonstrate the efficacy of the proposed framework. First, we will show the optimal solution of Algorithm \ref{Algorithm2} (ALG. \ref{Algorithm2}) and conduct a comparison of its performance with two baselines. The first one, named OPT, is the optimal solutions of problem \eqref{MEC_Opt} with exhaustive search for the optimal response $\boldsymbol{\theta}^*$. The second one is called Baseline that considers the worst response amongst the participating clients to attain local consensus $\theta_{\textrm{th}}$ accuracy  with an offered price. This is an inefficient scheme but still enables us to attain feasible solutions. Finally, we analyze the system performance by varying different parameters, and conduct a comparison of the incentive mechanism with the baseline and their corresponding utilities. In our analysis, the smaller values of local consensus are of specific interest as they reflect the effectiveness of FL.
		\begin{figure}[t!]
		\centering
		\includegraphics[width=3in]{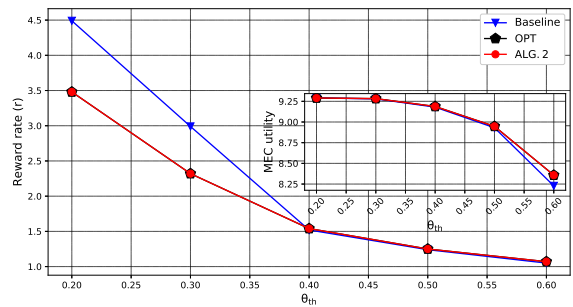}
		\caption{Comparison of (a) Reward rate and (b) MEC utility under three schemes for different values of threshold $\theta_{\textrm{th}}$ accuracy.}	
		\label{fig: Optimal_reward}
	\end{figure}

	\textit{1) Settings:}
	For an illustrative scenario, we fix the number of participating clients to 4. We consider the system parameter $\beta =10$, and the upper bound to the number of global iterations $\delta = 10,$ which characterizes the permissible rounds of communication to ensure global $\epsilon$ accuracy. The MEC's utility $U(x(\epsilon)) = 1 - 10^{-(ax(\epsilon)+b)}$ model is defined with parameters $a = 0.3$, and $b= 0$. For each client $k$, we consider normalized weight $\nu_k$ is uniformly distributed on [0.1,0.5], which can provide an insight on the system's efficacy as presented in Fig. \ref{fig:response_case_study}. We characterize the interaction between the MEC server and the participating clients under homogeneous channel condition, and use the normalized value of $T_k$ for all participating clients.
	\begin{figure*}[t!]
		\centering
		\begin{subfigure}[b]{0.3\textwidth}
			\centering
			\includegraphics[width=3in]{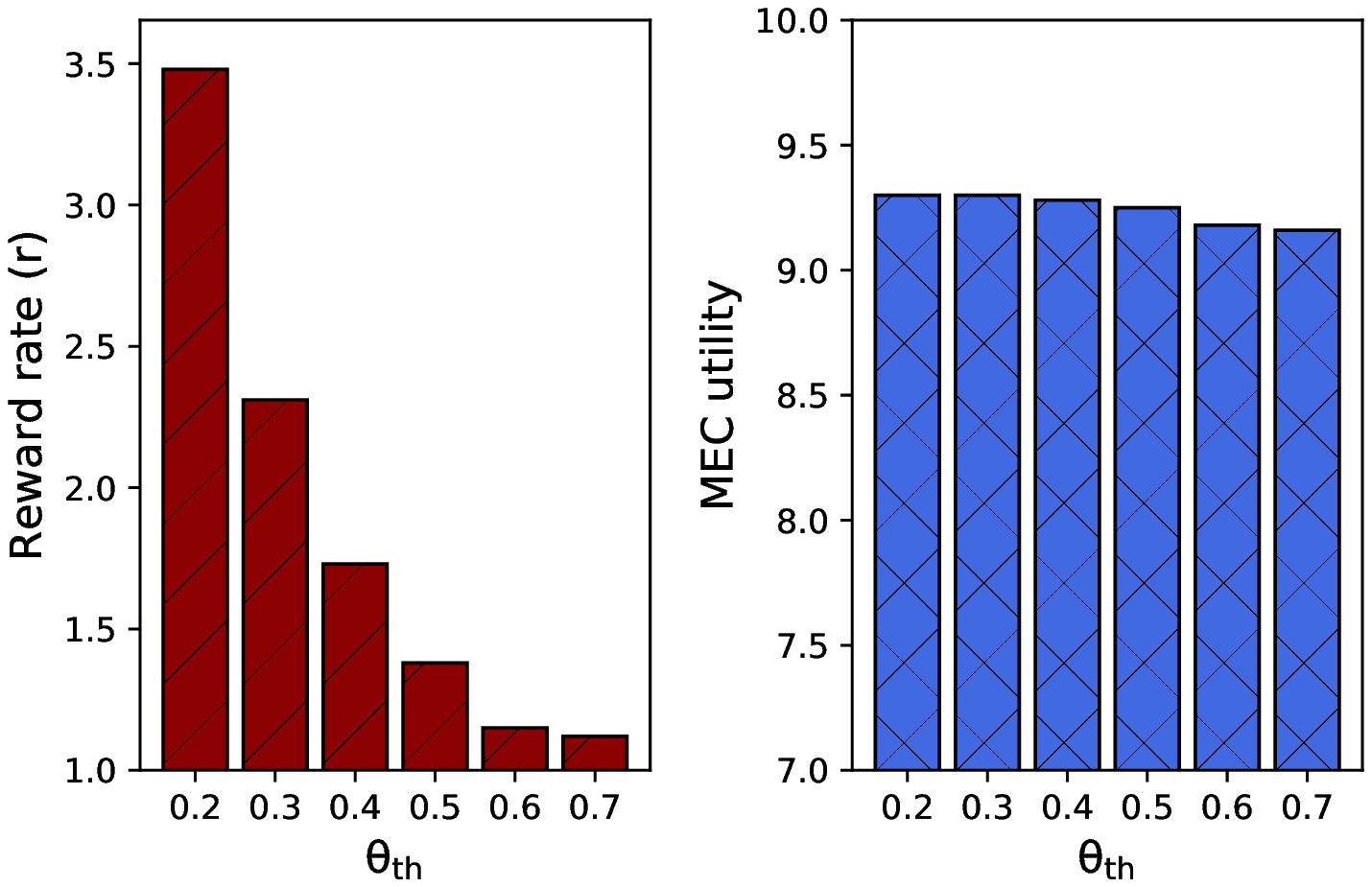}
			\caption{}
			\label{fig: simu_case1}
		\end{subfigure}%
		\hspace{3.5cm}
		\begin{subfigure}[b]{0.45\textwidth}
			\centering
			\includegraphics[width=3in]{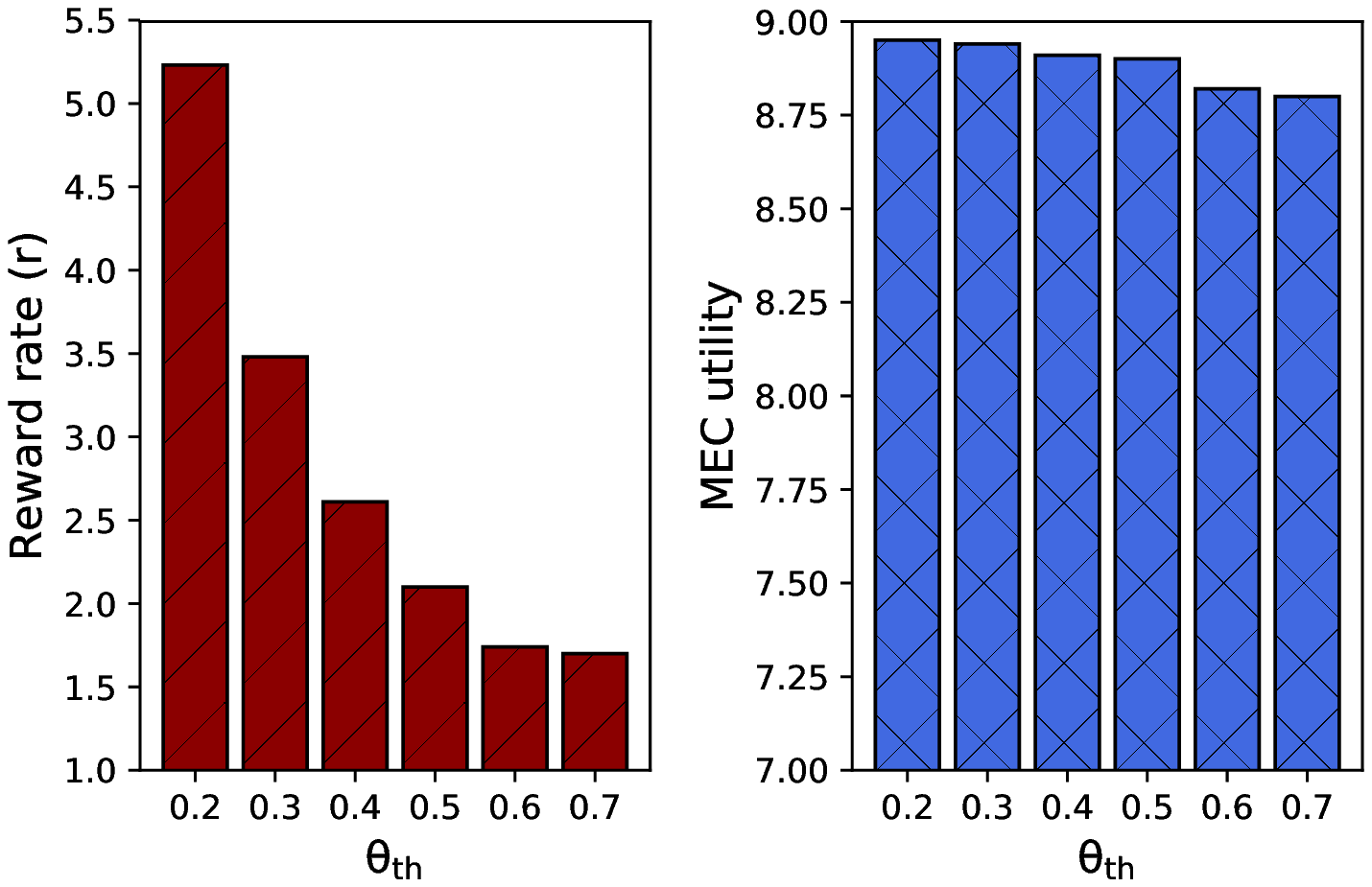}
			\caption{}
			\label{fig: simu_case2}
		\end{subfigure}
		\centering
		\caption{ (a) For $|\mathcal{K}| = 4, a= 0.3, b= 0$, $\gamma_k = 1, \forall k$. (b) For $|\mathcal{K}| = 4, a= 0.3, b= 0$, and $\gamma_k \sim U[1,5].$ }
		\label{fig:parametric_choice}
	\end{figure*}
	
	\textit{2) Reward rate:} In Fig. \ref{fig: Optimal_reward} we increase the value of local consensus accuracy $\theta_{\textrm{th}}$ from 0.2 to 0.6. When the  accuracy level is improved (from 0.4 to 0.2), we observe a significant increase in reward rate. These results are consistent with the analysis in Section \ref{sec:Incentive_Mechanism}-B. The reason is that cost for attaining higher local accuracy level requires more local iterations, and thus the participating clients exert more incentive to compensate for their costs. 
	
	We also show that the reward variation is prominent for lower values of $\theta_{\textrm{th}}$, and observe that scheme ALG. 2 and OPT achieve the same performance, while Baseline is not as efficient as others. Here, we can observe up to 22\%  gain in the offered reward against the Baseline by other two schemes. In Fig. \ref{fig: Optimal_reward}b, we see the corresponding MEC utilities for the offered reward that complements the competence of proposed ALG. 2. We see, the trend of utility against the offered reward goes along with our analysis.   
	
	\textit{3) Parametric choice:} In Fig. \ref{fig:parametric_choice} we show the impact of parametric choice adopted by the participating client $k$ to solve the local subproblem \cite{shamir2014distributed}, which is characterized by $\gamma_k$. In Fig. \ref{fig:parametric_choice}a, we see a lower offered reward for the improved local accuracy level for the participating clients adapting same parameters (algorithms) for solving the local subproblem, in contrast to Fig. \ref{fig:parametric_choice}b with the uniformly distributed $\gamma_k$ on [1,5] to achieve the competitive utility.
	
	\textit{4) Comparisons:} In Table a, and Table b, we see the effect of randomized parameter $\gamma_k$ for different configuration of MEC utility  model $U(\cdot)$ defined by $(a,b)$. For the smaller values of $\theta_{\textrm{th}}$, which captures the competence of the proposed mechanism, we  observe that the choice of $(a,b)$ provides a consistent offered reward for improved utility from $(0.35,-1)$ to $(0.65,-1)$, which follows our analysis in Section \ref{sec:Incentive_Mechanism}-A. For larger values of $\theta_{\textrm{th}}$, we also see the similar trend in MEC utility. For a randomized setting, we observe up to 71\% gain in offered reward against the Baseline, which validates our proposal's efficacy aiding FL. 
	
	\begin{table*}
		\centering
		
		\captionsetup[subtable]{position = below}
		\captionsetup[table]{position=top}
		\begin{subtable}{0.48\linewidth}
			\centering
			\addtolength{\tabcolsep}{-3pt}
			\begin{tabular}{|c|c|c|c|c|} \hline
				\textbf{Threshold accuracy} & \textbf{Baseline} & \textbf{ALG. 2} & \textbf{ALG. 2} & \textbf{ALG. 2} \\ 
				$\theta_{\textrm{th}}$ & $r$ & $(0.3,-1)$ & $(0.35,-1)$  & $ (0.65,-1)$ \\ 
				\hline
				0.2 & 18 & 5.22 & 5.22 & 5.22\\
				0.3 & 12 & 3.48 &3.48 & 3.48\\
				0.4 & 8.99 & 2.602 & 2.6 & 2.61\\
				0.5 & 7.19 & 2.79 & 4.3 & 2.2\\
				0.6 & 5.99 & 2.38 & 2.87 & 2.1\\
				0.7 & 5.13 & 2.84 & 3.17 & 1.9\\ 
				\hline
			\end{tabular}
			
			\caption{\small Offered reward rate comparison with randomized $\gamma$ effect for different $(a,b)$ setting.}
			\label{tab:table_i}
		\end{subtable}%
		\hspace*{2.9em}
		\begin{subtable}{0.48\linewidth}
			\centering
			\addtolength{\tabcolsep}{-3pt}
			\begin{tabular}{|c|c|c|c|}
				\hline
				\textbf{Threshold accuracy} & \textbf{ALG. 2} & \textbf{ALG. 2} & \textbf{ALG. 2} \\ 
				$\theta_{\textrm{th}}$ & $(0.3,-1)$ & $(0.35,-1)$  & $ (0.65,-1)$ \\ 
				\hline
				0.2  & 8.55 & 8.79 & 8.96\\ 
				0.3 & 8.41 &8.60 & 8.95\\ 
				0.4 & 8.33 & 8.58 & 8.94\\ 
				0.5 & 8.2 & 8.73 & 8.91\\ 
				0.6 & 8.18 & 8.4 & 8.91\\ 
				0.7 & 7.8 & 8.51 & 8.86\\ 
				\hline
			\end{tabular}
			\caption{\small Utility comparison with randomized $\gamma$ effect for different $(a,b)$ setting.}
			\label{tab:table_ii}
		\end{subtable}
	\end{table*}
	\section{Threshold Accuracy Estimation : An Admission Control Strategy}\label{sec:Admission_Control}

	Our earlier discussion in Section \ref{sec:Incentive_Mechanism} and simulation results explain the significance of choosing a local $\theta_{\textrm{th}}$ accuracy to build a global model that maximizes the utility of the MEC server. In this regard, at first, the MEC server evokes admission control to determine $\theta_{\textrm{th}}$ and the final model is learned later. This means, with the number of expected clients, it is crucial to appropriately select a proper prior value of $\theta_{\textrm{th}}$ that corresponds to the participating client's selection criteria for training a specific learning model. Note that, in each communication round of synchronous aggregation at the MEC server, the quality of local solution benefits to evaluate the performance at the local subproblem. In this section, we will discuss about the probabilistic model employed by the MEC server to determine the value of the consensus $\theta_{\textrm{th}}$ accuracy.
	
	We consider the local $\boldsymbol{\theta}$  accuracy  for the participating clients is an i.i.d and uniformly distributed random variable over the range $[ \theta_{\min}, \theta_{\max} ]$, then the PDF of the responses can be defined as $f_{\boldsymbol{\theta}}(\theta) = \frac{1}{\theta_{\max} - \theta_{\min}}$. Let us consider a sequence of discrete time slots $t \in \{1,2,\ldots\}$, where the MEC server updates its configuration for improving the accuracy of the system. Following our earlier definitions, at time slot $t$, the number of participating clients in the crowdsourcing framework for FL is $|\mathcal{K}(t)|$, or simply $K$. We restrict the clients with the accuracy measure $\boldsymbol{\theta}(t) \ge \theta_{\max}$. For $K$ number of participation requests, the total number of accepted responses $N(t)$ is defined as $N(t) = K \cdot F_{\boldsymbol{\theta}(t)}(\theta) = K \cdot P[ \boldsymbol{\theta}(t) \le \theta]$. We have $N(t) = K \cdot \left[\frac{\theta(t) - \theta_{\min}}{\theta_{\max} - \theta_{\min}} \right]$. At each time $t$, the MEC server chooses $\boldsymbol{\theta}(t)$ as the threshold accuracy $\theta_{th}$ that maximizes the sum of its utility as defined in (\ref{MEC_Utility}) for the defined parameters $ a \ge 0,b \le 0$ and the total participation,  $\beta \left(1 - 10^{-(ax(\epsilon)+b)} \right) + ( 1 - \theta ) \cdot N(t)$, subject to the constraint that the response lies between the minimum and maximum accuracy measure ($\theta_{\min} \le \boldsymbol{\theta}(t) \le \theta_{\max} $). Using the definitions in (\ref{CoCoA_bounded}), for $\beta > 0$, the MEC server maximizes its utility for the number of participation with $\theta$ accuracy as 
	\begin{align}
	\begin{split}
	\underset{\theta(t)}{\text{max}} \qquad & \beta \left(1 - 10^{-(a \cdot \delta (1 - \theta (t))+b)} \right) + (1-\theta (t)) \cdot N(t),
	\label{eq: participation_optimization}	\\
	\text{s.t.}\qquad & \theta_{\min} \le \theta(t) \le \theta_{\max}. 
	\end{split}
	\end{align}
		\begin{figure*}[t!]
		\centering
		\begin{subfigure}[b]{0.3\textwidth}
			\centering
			\includegraphics[width=3in]{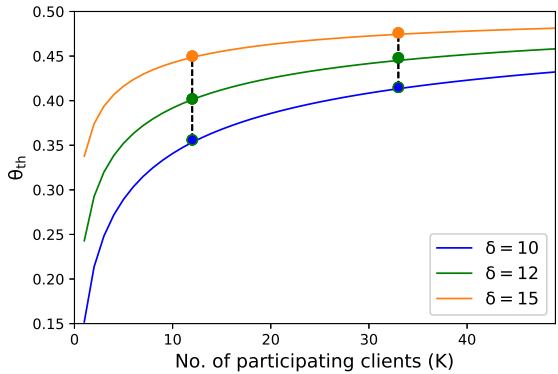}
			\caption{}
		\end{subfigure}%
		\hspace{3.5cm}
		\begin{subfigure}[b]{0.5\textwidth}
			\centering
			\includegraphics[width=3in]{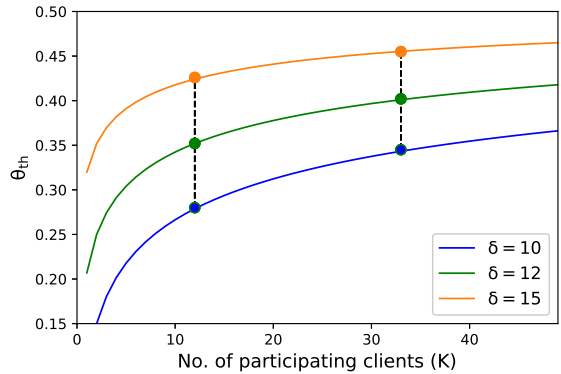}
			\caption{}
		\end{subfigure}
		\centering
		\caption{Variation of local $\theta_{\textrm{th}}$ accuracy for different values of $\delta$ given the density function, $f_{\boldsymbol{\theta}}(\theta) \sim U[0.1,0.9],|\mathcal{K}| = [0,50]$, (a) For a = 0.35, b = -1. (b) For a = 0.45, b = -1.05.}
		\label{fig:extension}
	\end{figure*}
	The Lagrangian of the problem (\ref{eq: participation_optimization}) is as follows:
	\begin{align}
	\mathcal{L}(\theta(t),\lambda, \mu) =\beta \left(1 - 10^{-(a \cdot \delta (1 - \theta (t))+b)} \right)  + (1-\theta(t)) \cdot \notag \\
	\left[\frac{\theta(t) - \theta_{\min}}{\theta_{\max} - \theta_{\min}} \right]  + \lambda (\theta(t) - \theta_{\min}) \notag
	\\+ \mu(\theta_{\max} - \theta(t)), 
	\end{align}
	where $\lambda \ge 0$ and $\mu \ge 0$ are dual variables. Problem (\ref{eq: participation_optimization}) is a convex problem whose optimal primal and dual variables can be characterized using the Karush-Khun-Tucker (KKT) conditions \cite{boyd2004convex} as 
	\begin{align}
	\frac{\partial \mathcal{L}}{\partial \theta(t)} = \ln(10) \cdot (\beta \delta a) \cdot 10^{-(a \cdot \delta (1 - \theta^* (t))+b)} \notag\\- K \cdot \left[\frac{2 \theta(t) - \theta_{\min}}{\theta_{\max} - \theta_{\min}}\right] + \lambda - \mu  = 0, \label{kkt:prob}\\
	\lambda(\theta(t) - \theta_{\min}) = 0, \label{kkt:cs1}\\
	\nu (\theta_{\max} - \theta(t)) = 0.
	\end{align}
	Following the complementary slackness criterion, we have 
	\begin{equation}
	\lambda^*(\theta^*(t) - \theta_{\min}) = 0 , \mu^* (\theta_{\max} - \theta^*(t)) = 0, \lambda^* \ge 0,\mu^* \ge 0. \label{kkt:cs}
	\end{equation}
	Therefore, from \eqref{kkt:cs}, we solve (\ref{eq: participation_optimization}) with the KKT conditions assuming that $\theta^*(t) < \theta_{\textrm{max}} $ as an admission control strategy, and find the optimal $\theta^*(t)$ that satisfies the following relation 
	
	\begin{equation}
	\begin{aligned}
	K & = \frac{\ln(10) \cdot (\beta \delta a) \cdot 10^{-(a \cdot \delta (1 - \theta^* (t))+b)} \cdot (\theta_{\min} - \theta_{\max})}{1 - 2 \theta^*(t) + \theta_{\min}}. 
	\label{eq: participation_threshold_theta}
	\end{aligned}
	\end{equation}
	
	(\ref{eq: participation_threshold_theta}) can be rearranged as 
	\begin{align}
	f(\theta^*(t)) = \ln(10) \cdot (\beta \delta a) \cdot 10^{-(a \cdot \delta (1 - \theta^* (t))+b)} \notag \\
	+ K\cdot  \left[ \frac{1 - 2 \theta^*(t) + \theta_{\min}}{\theta_{\max} - \theta_{\min}}  \right] =0 . 
	\end{align}
	
	To obtain the value of $\theta^*(t)$ we will use \textit{Netwon-Raphson method} \cite{conte2017elementary} employing an appropriate initial guess that manifests the quadratic convergence of the solution. We choose $\theta^*_0 (t) = E(\boldsymbol{\theta}(t)) = \frac{\theta_{\max}+\theta_{\min}}{2}$ as an initial guess for finding $\theta^*(t)$ which follows the PDF $f_{\boldsymbol{\theta}}(\theta)  \sim U[\theta_{\min},\theta_{\max}]$. Then the solution method is an iterative approach as follows:
	\begin{align}
	\begin{split}
	\theta_{i+1}^*(t) &=  \theta^*_{i}(t) - \frac{f(\theta_{i}^*(t))}{\beta\delta^2 a^2 \cdot \ln^2(10) \cdot 10^{-(a \cdot \delta (1 - \theta_i^* (t))+b)}}. 
	\end{split} 
	\end{align}

	\textit{Numerical Analysis:} In Fig. \ref{fig:extension}, we vary the number of participating clients up to 50 with different values of $\delta$. The response of the clients is set to follow a uniform distribution on [0.1, 0.9] for the ease of representation. In Fig. \ref{fig:extension}a, for the model parameters (a,b) as (0.35,-1), we see $\theta_{\textrm{th}}$ increases with the increase in the number of participating clients for all values of $\delta$. It is intuitive, and goes along with our earlier analysis that for the small number of participating clients, the smaller $\theta_{\textrm{th}}$ captures the efficacy of our proposed framework. Because it is an iterative process, the evolution of $\theta_{\textrm{th}}$ over the rounds of communication will be reflected in the framework design. Subsequently, the larger upper bound $\delta$ exhibits the similar impact on setting $\theta_{\textrm{th}}$, where smaller $\delta$ imposes strict local accuracy level to attain high-quality centralized model. Also due to the same reason, in Fig. \ref{fig:extension}b, we see $\theta_{\textrm{th}}$ is increasing for the increase in the number of participating clients, however, with the lower value. It is because of the choice of parameters (a,b) as explained in Section \ref{sec:Incentive_Mechanism}-A. So the value of $\theta_{\textrm{th}}$ is lower in Fig. \ref{fig:extension}b.

	\section{Conclusion} \label{sec:Conclusions}

	In this paper, we have designed and analyzed a novel crowdsourcing framework to enable FL. An incentive mechanism has been established to enable the participation of several devices in FL. In particular, we have adopted a two-stage Stackelberg game model to jointly study the utility maximization of the participating clients and MEC server interacting via an application platform for building a high-quality learning model. We have incorporated the challenge of maintaining communication efficiency for exchanging the model parameters among participating clients during aggregation. Further, we have derived the best response solution and proved the existence of Stackelberg equilibrium. We have examined characteristics of participating clients for different parametric configurations. Additionally, we have conducted numerical simulations and presented several case studies to evaluate the framework efficacy. Through a probabilistic model, we have designed and presented numerical results on an admission control strategy for the number of client's participation to attain the corresponding local consensus accuracy. For future work, we will focus on mobile crowdsourcing framework to enable the self-organizing FL that considers task offloading strategies for the resource constraint devices. We will consider the scenario where the central coordinating MEC server is replaced by one of the participating clients and devices can offload their training task to the edge computing infrastructure. Another direction is to study the impact of discriminatory pricing scheme for participation. Such works can narrate towards numerous incentive mechanism design, such as offered tokens in blockchain network \cite{kim2018device}. We also plan to further investigate on participating client's behavior, in terms of incentive and communication efficiency, to incorporate cooperative data trading scenario for the proposed framework \cite{yu2017mobile}, \cite{jiao2018profit}.

	\begin{appendices}
		\section{KKT Solution}
		The utility maximization problem in (\ref{eq:MEC_Utility}) is a convex optimization problem whose optimal solution can be obtained by using Lagrangian duality. The lagrangian of (\ref{eq:MEC_Utility}) is 
		\begin{align*}
		\begin{split}
		\mathcal{L}(r,x(\epsilon),\lambda) &= \beta \left(1 - 10^{-(ax(\epsilon)+b)} \right) - r \sum_{k \in K}(1 - \theta^*_k(r)) \notag\\
		& +\lambda \left[  \delta(1 - {\max}_k \ \theta^*_k(r) - x(\epsilon) \right]	    
		\end{split}    \tag{A.1}   \label{eq:Lagrangian} 
		\end{align*}
		where $\lambda \geq 0$ is the Lagrangian multiplier for constraint (\ref{eq : constraint}).
		\\
		By taking the first-order derivative of (\ref{eq:Lagrangian}) with respect to $x(\epsilon)$ and $\lambda$, KKT conditions are expressed as follows:	
		\begin{align*}
		\frac{\partial\mathcal{L}}{\partial x(\epsilon)} &= a \beta e^{-(a(x(\epsilon)) + b)} - \lambda \le 0, \text{\ if } x(\epsilon) \ge 0.
		\notag     \tag{A.2} \\     
		\frac{\partial\mathcal{L}}{\partial \lambda} &=  \left[  \delta(1 - {\max}_k \ \theta^*_k(r)) - x(\epsilon) \right] \ge 0 , \text{\ if } \lambda \ge 0.   \tag{A.3}
		\end{align*}
		By solving (A.2), the solution to the utility maximization problem (\ref{eq:MEC_Utility}) is 	
		\begin{equation}
		x^*(\epsilon) = \frac{-\ln(\lambda/a\beta)-b}{a}.    \tag{A.4} 
		\end{equation}      
		From (A.3), the Lagrangian multiplier $\lambda$ is as 
		\begin{equation}
		\lambda^* = a \beta  e^{[a\delta ( 1 - {\max}_k \ \theta^*_k(r))+ b ]}.  \tag{A.5}
		\end{equation}
		Thus, from (A.4) and (A.5) the optimal solution to the utility maximization problem (\ref{eq:MEC_Utility}) is 
		\begin{equation}
		x^*(\epsilon) = \delta ( 1 - {\max}_k \ \theta^*_k(r)). \tag{A.6}
		\end{equation}
	\end{appendices}
	\bibliographystyle{IEEEtran}
	\bibliography{mybib}
\end{document}